\documentclass{article}

\usepackage[utf8]{inputenc} % allow utf-8 input
\usepackage[T1]{fontenc}    % use 8-bit T1 fonts
\usepackage[svgnames,usenames,dvipsnames,table]{xcolor}         % colors
\usepackage[colorlinks=true, citecolor=Navy, breaklinks=true]{hyperref}       % hyperlinks
\usepackage{url}            % simple URL typesetting
\usepackage{booktabs}       % for professional tables
\usepackage{amsfonts}       % blackboard math symbols
\usepackage{nicefrac}       % compact symbols for 1/2, etc.
\usepackage{microtype}

\usepackage{algorithm}
\usepackage{algorithmic}

     \PassOptionsToPackage{numbers, compress}{natbib}
\usepackage{cereb}
\usepackage{environ}

\usepackage{amsmath}
\usepackage{amssymb}
\usepackage{mathtools}
\usepackage{amsthm}

\usepackage[capitalize,noabbrev]{cleveref}

\usepackage{wrapfig}
\usepackage{graphicx}
\usepackage{subcaption}
\usepackage{multirow}
\usepackage{tcolorbox}
\usepackage{paralist}

\theoremstyle{plain}
\newtheorem{theorem}{Theorem}[section]

\theoremstyle{definition}
\newtheorem{definition}[theorem]{Definition}

\theoremstyle{remark}

\usepackage[disable,textsize=tiny]{todonotes}
\usepackage{enumitem}
\setlist[itemize]{left=0.5em}
\setlist[enumerate]{left=0.5em}

\crefname{figure}{Fig.}{Figs.}
\Crefname{figure}{Fig.}{Figs.}
\crefname{equation}{Eq.}{Eqs.}
\Crefname{equation}{Eq.}{Eqs.}
\crefname{section}{Sec.}{Secs.}
\Crefname{section}{Sec.}{Secs.}
\crefname{subsection}{Sec.}{Secs.}
\Crefname{subsection}{Sec.}{Secs.}
\crefname{subsubsection}{Sec.}{Secs.}
\Crefname{subsubsection}{Sec.}{Secs.}
\crefname{theorem}{Theorem}{Theorems}
\Crefname{theorem}{Theorem}{Theorems}
\crefname{lemma}{Lemma}{Lemmas}
\Crefname{lemma}{Lemma}{Lemmas}
\crefname{proposition}{Proposition}{Propositions}
\Crefname{proposition}{Proposition}{Propositions}
\crefname{corollary}{Corollary}{Corollaries}
\Crefname{corollary}{Corollary}{Corollaries}
\crefname{definition}{Definition}{Definitions}
\Crefname{definition}{Definition}{Definitions}
\crefname{assumption}{Assumption}{Assumptions}
\Crefname{assumption}{Assumption}{Assumptions}
\crefname{remark}{Remark}{Remarks}
\Crefname{remark}{Remark}{Remarks}
\crefname{example}{Example}{Examples}
\Crefname{example}{Example}{Examples}

\newcommand{\appendixwidth}{0.4}  % Handy to say \flops in equations

\newcommand{\mup}{\mbox{$\mu$P}}

\newcommand{\maxlr}{1.6\mbox{e-}02}
\newcommand{\maxlrdetail}{1.62\mbox{e-}02}

\DeclareMathOperator*{\argmin}{arg\,min}

\newcommand{\lhat}{\hat{L}}
\newcommand{\hateta}{\tilde{\eta}}
\newcommand{\cbs}{B_{\text{crit}}}
\newcommand{\bcrit}{\cbs}  % because why not have two?
\newcommand{\cbszhang}{B_{\text{zhang}}}
\newcommand{\titer}{\tau_\text{iter}}
\newcommand{\tepochwang}{\tau_\text{epoch}}
\newcommand{\tepoch}{\tau}
\newcommand{\tema}{\tepoch}  % twice as nice
\newcommand{\dmin}{D_{\text{min}}}
\newcommand{\smin}{S_{\text{min}}}

\newcommand{\nconst}{N_{\text{const}}}
\newcommand{\dconst}{D_{\text{const}}}

\newcommand{\dinfer}{D_B}
\newcommand{\irr}{E_N}  % irreducible loss including N
\newcommand{\hatr}{\hat{r}}
\newcommand{\kconst}{K_{\text{const}}}

\newcommand{\bopt}{B_{\text{opt}}}
\newcommand{\boptcond}{B_{\text{opt}|\lambda}}
\newcommand{\bdeepseek}{B_{\text{deepseek}}}
\newcommand{\tepochopt}{{\tepoch}_{\text{opt}}}
\newcommand{\temaopt}{\tepochopt}
\newcommand{\lambdaopt}{\lambda_{\text{opt}}}
\newcommand{\etaopt}{\eta_{\text{opt}}}
\newcommand{\hatetaopt}{\tilde{\eta}_{\text{opt}}}
\newcommand{\nopt}{N_{\text{opt}}}
\newcommand{\dopt}{D_{\text{opt}}}
\newcommand{\varopt}{\sigma^2_{\text{opt}}}
\newcommand{\cetad}{c_{\eta_D}}
\newcommand{\metad}{m_{\eta_D}}
\newcommand{\flops}{\text{FLOPs}}  % Handy to say \flops in equations
\newcommand{\totflopstext}{\text{Total $\flops$}}
\newcommand{\minflops}{C(N, \dmin)}
\newcommand{\totflops}{C_{\text{+}}(N,\dmin,B)}
\newcommand{\hatf}{\hat{C}}
\newcommand{\ttime}{\text{Training Time}}

\newcommand{\leftfig}{\text{\emph{left}}}
\newcommand{\middlefig}{\text{\emph{middle}}}
\newcommand{\rightfig}{\text{\emph{right}}}

\newcommand{\cbsPoints}{\mathit{bcrit\_scaling\_law\_fitting\_points}}
\newcommand{\scaleSet}{\mathit{scaling\_law\_fitting\_points}}
\newcommand{\tradeoffSet}{\mathit{tradeoff\_fitting\_points}}
\newcommand{\modelScales}{\mathit{model\_scales}}
\newcommand{\batchSizes}{\mathit{batch\_sizes}}

\newcommand{\LLM}{\mathit{LLM}}
\newcommand{\ccbsalg}{c}
\newcommand{\mcbsalg}{m}
\newcommand{\ctepochalg}{c}
\newcommand{\mtepochalg}{m}
\newcommand{\tepochPoints}{\mathit{tau\_scaling\_law\_fitting\_points}}
\newcommand{\lossPoints}{\mathit{loss\_points}}
\newcommand{\lambdaRange}{\mathit{lambda\_range}}
\newcommand{\tpp}{\mathrm{TPP}}
\newcommand{\ccbs}{c_{\cbs}}
\newcommand{\mcbs}{m_{\cbs}}
\newcommand{\ctepoch}{c_{\tepoch}}
\newcommand{\mtepoch}{m_{\tepoch}}

\newcommand{\mbopt}{m_{\bopt}}
\newcommand{\Rtwo}{R^{2}}

\makeatletter
\@ifundefined{ack}{
  \NewEnviron{ack}{
    \section*{Acknowledgments and Disclosure of Funding}
    \BODY
  }
}{}
\makeatother

\newcounter{fcounter}
\newcommand\finding[1]{
        \refstepcounter{fcounter}\vspace{2pt}
        \begin{tcolorbox}[colback=yellow!10!white,colframe=yellow!80!black,boxsep=1pt,left=2pt,right=2pt,top=1pt,bottom=1pt]\noindent{\textbf{\sffamily Finding \arabic{fcounter}}: \sffamily #1}
        \end{tcolorbox}\vspace{0pt}
}
\crefname{fcounter}{Finding}{Findings}

\newcounter{kcounter}
\newcommand\takeaway[1]{
        \refstepcounter{kcounter}\vspace{2pt}
        \begin{tcolorbox}[colback=green!10!white,colframe=green!80!black,boxsep=1pt,left=2pt,right=2pt,top=1pt,bottom=1pt]\noindent{\textbf{\sffamily Key takeaway \arabic{kcounter}}: \sffamily #1}
        \end{tcolorbox}\vspace{0pt}
}
\crefname{kcounter}{Takeaway}{Takeaways}

\renewcommand{\cite}[1]{\PackageError{MyPackage}{Do not use \string\cite\space with natbib. Use \string\citet\space or \string\citep}{See the natbib package documentation for explanation.}}

\makeatletter
\AtBeginDocument{%
  \@ifpackageloaded{cleveref}{%

    \providecommand\cref@appendix@setup{%
      \crefname{appendix}{Appendix}{Appendices}%
      \Crefname{appendix}{Appendix}{Appendices}%
    }%

    \AddToHook{cmd/appendix/before}{%
      \cref@appendix@setup
      \@ifundefined{chapter}{%
        \crefalias{section}{appendix}%
      }{%
        \crefalias{chapter}{appendix}%
      }%
      \crefalias{subsection}{appendix}%
      \crefalias{subsubsection}{appendix}%
    }%

  }{}% end \@ifpackageloaded{cleveref}
}
\makeatother

\title{Power Lines: Scaling Laws for Weight Decay and Batch Size in LLM Pre-training}

\author{Shane Bergsma, Nolan Dey, Gurpreet Gosal, Gavia Gray, Daria Soboleva, Joel Hestness \\
  Cerebras Systems \\
  \texttt{\{shane.bergsma,joel\}@cerebras.net}
}

\begin{document}

\maketitle

\begin{abstract}
Efficient LLM pre-training requires well-tuned hyperparameters (HPs),
including learning rate $\eta$ and weight decay $\lambda$.
We study \emph{scaling laws} for HPs: formulas for how to scale HPs as
we scale model size $N$, dataset size $D$, and batch size $B$.
Recent work~\citep{wang2024how} suggests the AdamW timescale, $\tema =
B/(\eta\lambda D)$, should remain constant across training settings,
and we verify the implication that optimal $\lambda$ scales linearly
with $B$, for a \emph{fixed} $N$ and $D$.  However, as $N$ and
$D$ \emph{scale}, we show optimal $\tema$ obeys a precise power law
%--- a relationship that appears as a straight line on a log-log plot ---
in the tokens-per-parameter ratio, $D/N$.  This law thus provides a
method to accurately predict $\lambdaopt$ in advance of large-scale
training.
We also study scaling laws for optimal batch size $\bopt$ (the $B$
enabling lowest loss at a given $N, D$) and critical batch size $\bcrit$ (the
$B$ beyond which further data parallelism becomes ineffective).
In contrast to prior work, we find both $\bopt$ and $\bcrit$ scale as
power laws in $D$, independent of model size, $N$.
Finally, we analyze how these findings inform the real-world selection
of Pareto-optimal $N$ and $D$ under dual training time and compute
objectives.
All experiments were run on Cerebras CS-3 systems.

\end{abstract}

\begin{figure}[h]
    \centering
    \begin{minipage}{0.33\textwidth}
        \includegraphics[trim={0.3cm 0.4cm 0.264cm 0.3cm}, clip, width=\linewidth]{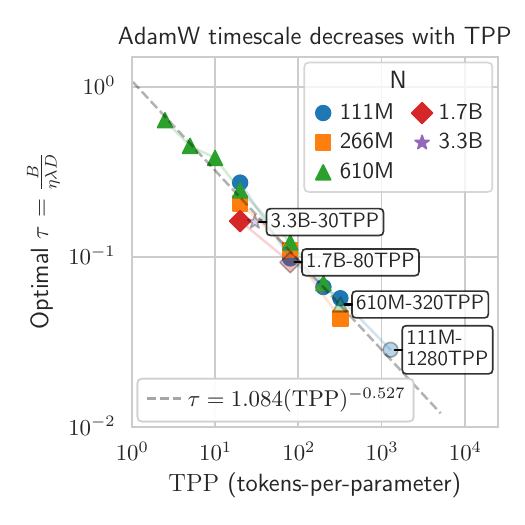}
    \end{minipage}\hfill
    \begin{minipage}{0.33\textwidth}
        \includegraphics[trim={0.3cm 0.4cm 0.264cm 0.3cm}, clip, width=\linewidth]{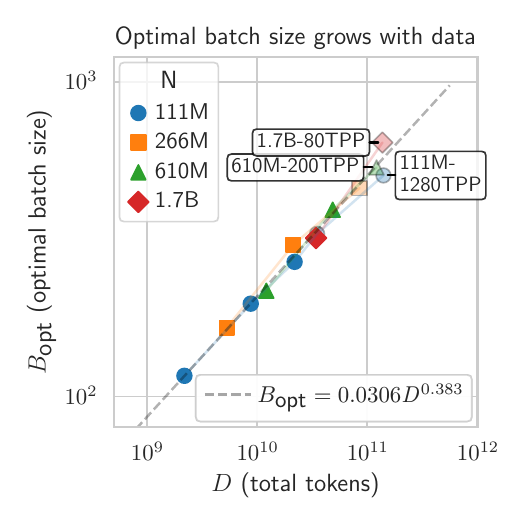}
    \end{minipage}\hfill
    \begin{minipage}{0.33\textwidth}
        \includegraphics[trim={0.3cm 0.4cm 0.264cm 0.3cm}, clip, width=\linewidth]{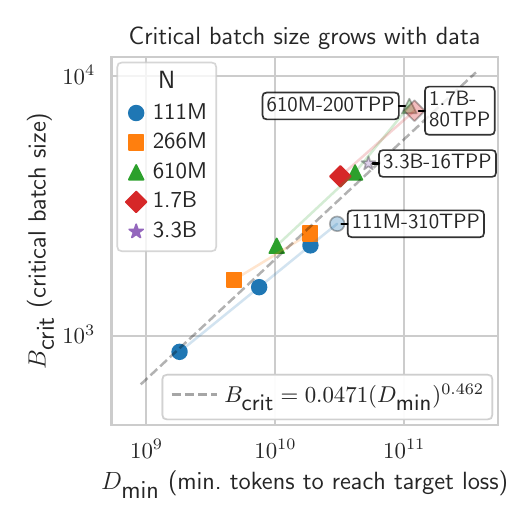}
    \end{minipage}
    \caption{\textbf{Hyperparameters and their power lines}: Optimal
      $\tepoch$ obeys a power law in tokens-per-parameter
      ($\leftfig$), while optimal batch size ($\middlefig$) and
      critical batch size ($\rightfig$) obey power laws in
      $D$. \textcolor[rgb]{0.5,0.5,0.5}{Faded} markers indicate points
      not used in fitting; all fits generalize well to larger-scale
      runs.
      \label{fig:hook}}
\end{figure}

\section{Introduction}\label{sec:intro}
LLMs predictably improve as model size $N$ and training data size $D$
increase~\citep{hestness2017scalinglaws,hoffmann2022empirical,kaplan2020scaling}.
Today, state-of-the-art LLMs are trained at computational scales that
leave no scope for hyperparameter (HP) tuning, although it is widely
accepted that good HPs are critical for effective
training~\citep{yang2022mup,wortsman2023small,bi2024deepseek}.

Both theoretical and empirical efforts have sought to address this.
Theoretically, maximal update parameterization ($\mup$) allows the
optimal learning rate $\etaopt$ and initial weight variance $\varopt$
to remain stable when scaling model
width~\citep{yang2020feature,yang2022mup},
% and depth~\citep{bordelon2023depthwise,yang2023tensor,dey2025dont},
enabling a ``tune small and train large'' strategy.
Empirically, DeepSeek~LLM~\citep{bi2024deepseek} adopted ``scaling
laws for HPs,'' where optimal batch size $\bopt$ and optimal learning
rate $\etaopt$ are estimated at small scale, and then extrapolated via
a power law fit in total compute FLOPs, $C$. A similar approach was
used in~\citet{kaplan2020scaling}, forecasting $\bopt$ and $\etaopt$
from loss $L$ and model size $N$.

Relying on a unique predicted $\bopt$ is \emph{inflexible}---it
precludes adjusting $B$ for compute/time trade-offs or hardware
constraints.
It is also unclear whether $C$, $L$, $N$, or $D$ (or a combination)
best explains scaling.  \citet{bi2024deepseek} noted, ``for models with
the same [$C$] but different model/data allocations, the optimal
parameter space varies slightly.''
Also, no comparable study has been done for weight decay $\lambda$.

This paper introduces a flexible, unified approach to HPs, using both
$\mup$ and scaling laws.
We fit power laws to losses derived from hundreds of $\mup$-trained
models, focusing on combinations of $\lambda$, $B$, $N$, and $D$.
We study both \emph{compute-optimal} and \emph{overtrained} models.
The fewest FLOPs to achieve a loss typically occurs when training at
$\approx$20 tokens-per-parameter (TPP =
$D/N$)~\citep{hoffmann2022empirical,besiroglu2024chinchilla}, but
overtrained models ($>$20~TPP) offer more-efficient
inference~\citep{touvron2023llama}.  We study TPPs from 20 to 1280.

To capture $\lambda$'s interaction with other scaling HPs ($\eta$,
$B$), we model scaling of the AdamW timescale, $\tepoch =
B/(\eta\lambda D)$.
\citet{wang2024how} found optimal $\tepoch$ stable with
varying $D$, but we show it obeys a power law in TPP (\cref{fig:hook},
$\leftfig$).
This law thus enables accurate estimation of $\lambdaopt$ for any $N$,
$D$, $B$.

Leveraging these better HPs as $B$ scales, we also study optimal batch
size: the $\bopt$ that minimizes loss at a given $N$ and $D$.
While $\bopt$ scales as a power law in $C$ \emph{when TPP is fixed}
(\cref{fig:related}, $\leftfig$), our results show this arises from a
more fundamental power-law dependence on $D$ (\cref{fig:hook},
$\middlefig$).

Importantly, increasing $B > \bopt$ can still reduce training time
(fewer steps) and improve hardware utilization.
This raises the question: how much \emph{extra} data is needed when
using large $B$?
Prior work defines the \emph{critical batch size} $\cbs$ as the point
where training to a target loss requires $2 \times \dmin$, with
rapidly-diminishing returns in training speed
thereafter~\citep{mccandlish2018empirical}.
We show $\cbs$ also scales with $D$ (\cref{fig:hook}, $\rightfig$),
not $L$ as suggested in~\citet{kaplan2020scaling} (\cref{fig:related},
$\middlefig$), consistent with recent results
from~\citet{zhang2024how}.

Finally, amid intense competition to advance LLM performance, a key
question is: which $N$, $D$, and $B$ yield the best trade-off between
training speed and compute cost?
Using our fit $\cbs$ law, we derive Pareto-optimal solutions to these
competing objectives, and show that small, overtrained models can be
best---offering both faster steps and greater parallelism via larger
$D$ (and thus higher $\cbs$).

Key findings and takeaways are highlighted in the paper. Our main
contributions are:
\begin{itemize}[topsep=1pt,itemsep=1pt]
\item The first large-scale empirical study varying weight decay $\lambda$ across $N$, $D$, and $B$ in LLMs.
\item Showing the AdamW timescale obeys a power law, enabling $\lambdaopt$ for any $N$, $D$, $B$ (\cref{sec:lambda}).
\item A new method for estimating $\cbs$, suitable for any LR schedule or optimizer (\cref{sec:cbs_methods}).
\item Confirmation that both $\bopt$ and $\cbs$ scale as power laws in $D$ (\cref{sec:bcrit}).
\item New methods for selecting $N$, $D$, and $B$ to trade-off training time vs.\@ compute (\cref{sec:balancing}).
\end{itemize}

\section{Scaling of the AdamW timescale $\tepoch$, and optimal weight decay $\lambdaopt$}\label{sec:lambda}
%Ideally, we could choose $N$, $D$, and $B$ based on time or compute
%constraints, and be confident that the resulting models would be
%well-tuned.

%Good HPs enable good training, and also good scaling laws.

%%
%Before we could reliably measure $B$ scaling, we found AdamW HPs to be
%suboptimal under $\mup$ as $B$, $D$ varied.
%%
%This led to our study of optimal HPs, culminating in the $\tepoch$
%scaling law.

\subsection{Background: $\mup$, AdamW, and $\tepochwang$}

\paragraph{$\mup$}\label{sec:mup}
$\mup$ is increasingly used in LLM
training~\citep{dey2023cerebras,dey2023btlm3b8k,sengupta2023jais,shen2024power,hu2024minicpm,abdin2024phi}.
With $\mup$, base HPs are tuned on a \emph{proxy} model and then
transferred to wider~\citep{yang2022mup} and
deeper~\citep{dey2025dont} models.
%with scaling factors for learning rate, initial weight variance, and
%other HPs dependent on the relative width of the proxy and target
%models.
Given the width of the proxy model, $d_p$, and target, $d_t$, $\mup$
prescribes scaling factors to apply to the LR, initial weight
variance, and other base HPs.
In particular, the optimal base LR, $\hatetaopt$ is scaled down to
$\etaopt = (\nicefrac{d_p}{d_t})\hatetaopt$.
%
%$\mup$ also uses scaling to stabilize embeddings, layer norms, and
%self-attention in transformers.
%
%Might reference \citet{dey2023cerebras} for how they adjust $B$ for
%MUP\@.  Note they don't say how to estimate $\cbs$.

While $\mup$ enables the same base LR to be used across different $N$,
$\hatetaopt$ has empirically been found to \emph{vary with
$B$}~\citep{yang2022mup,lingle2024large,noci2024learning,shen2024power}.
Moreover, recent work has also observed $\hatetaopt$ \emph{decreasing
in $D$}, leading to proposals for scaling $\hatetaopt$ as a
(decreasing) power law in $D$~\citep{shen2024power,bjorck2024scaling}.
%
%In our experiments with AdamW, when $B$ and $D$ change, we
%consistently find it more effective to adjust $\lambda$ rather than
%$\eta$.

\paragraph{The EMA view of AdamW}\label{sec:ema}

Rather than adjusting $\eta$ as $D$ scales, \citet{wang2024how}
proposed that, if using the AdamW
optimizer~\citep{loshchilov2017decoupled} with $\mup$, then the weight
decay, $\lambda$, should instead be adjusted.
To see this, note an AdamW update at each step, $t$, can be expressed
in terms of $\eta \lambda$ as:
\begin{equation}\label{eq:adamw}
\theta_t = (1 - \eta\lambda)\theta_{t-1} - \eta \frac{\hat{m}_t}{\sqrt{\hat{v}_t} + \epsilon}
\end{equation}
Here, $\eta$ is the $\mup$-adjusted LR, and $\hat{m}_t$ and
$\hat{v}_t$ are (bias-corrected) exponentially-weighted moving
averages (EMAs) of gradients and squared
gradients~\citep{kingma2014adam}. \citet{wang2024how} observed AdamW's
parameters, $\theta_t$, can \emph{also} be viewed as an EMA---of
weight \emph{updates}.  Specifically, the standard EMA form \mbox{$y_t
  = (1 - \alpha)y_{t-1} + \alpha x_t$} matches AdamW when
%\begin{equation}\label{eq:adamwema}
\mbox{$y_t=\theta_t$}, \mbox{$\alpha=\eta\lambda$}, and
\mbox{$x_t=-\frac{1}{\lambda}\frac{\hat{m}_t}{\sqrt{\hat{v}_t} + \epsilon}$}.
%\end{equation}
The quantity $\nicefrac{1}{\alpha} = \nicefrac{1}{\eta\lambda}$
provides a measure of the number of iterations (i.e., steps) over
which updates are averaged;
\citet{wang2024how} denote it as $\titer$.
They show that if the timescale is measured in \emph{epochs} as
$\tepochwang = {\titer}/{M}$, where $M$ is iterations-per-epoch, then
the optimal $\tepochwang$ remains stable when $N$ or $D$ scale (on
image tasks).  I.e., if $M$ scales up, $\lambda$ should be
scaled \emph{down} to maintain constant $\tepochwang$.
%i.e., $M$ increases, $\lambda$ should be scaled down accordingly.
%
%they did not experiment with compute-optimal LLMs, but mainly image
%recognition datasets).
%
%%
%they recommend keeping
%$\tepochwang$ constant via a corresponding adjustment to
%$\lambda$.
%
%We expand on this approach below and show how to successfully use it
%when both $D$ \emph{and} $B$ vary.

\subsection{Methods: The AdamW timescale for LLMs, $\tepoch$, and its scaling}\label{sec:tema_methods}

%\paragraph{$\lambda$ scales with $B$}

Since LLM pre-training only uses one ``epoch'' of data, we normalize
the timescale as $\tepoch = {\titer}/{S}$, where $S$ is the total
number of optimization steps.\footnote{Because it depends on $\eta$,
timescale $\tepoch$ \emph{varies} during LR decay. Here we compute
$\tepoch$ at peak LR\@. Notably, if two training runs share the same
LR schedule (shape) and $\tepoch$ (at peak LR), their final AdamW
timescale (EMA contributions over the \emph{data}) will match---even
if $B$ differs. \cref{sec:ema_lr} provides further details.}
Moreover, since $S$=${D}/{B}$,
%then $\tepoch = \titer(\nicefrac{B}{D})$, i.e.,
%Given $\titer = \nicefrac{1}{\eta\lambda}$,
\begin{equation}\label{eq:tepoch}
  \tepoch = \frac{B}{\eta\lambda D}
\end{equation}

$\tepoch$ reflects the fraction of past iterations to include in the
final weights.
While \citet{wang2024how} did not vary $B$, their work suggests this
fraction should remain constant as $B$ scales.
We hypothesize that when moving from compute-efficient to overtrained
LLMs, updates can be integrated over a smaller fraction of the data;
specifically, that $\tepochopt$ decreases as a power law in $\tpp :=
\nicefrac{D}{N}$:
\begin{equation}\label{eq:tepoch_scaling}
  \tepochopt(\tpp) = \ctepoch \cdot \tpp^{\mtepoch}
\end{equation}
where $\ctepoch$ and $\mtepoch$ are parameters to be fit.
%
%In the following results, we demonstrate excellent fits to
%\cref{eq:tepoch_scaling} on our experimental data.
%
Appendix \cref{alg:tepoch_fitting} summarizes the fitting procedure.
%
%Note we are collecting $\tepoch$ here for runs where $B < \cbs$.
%
Taking the $\mup$-adjusted $\eta$ as our LR, $\lambdaopt$ can be
computed from \cref{eq:tepoch,eq:tepoch_scaling}:
\begin{equation}\label{eq:lambdaopt}
   \lambdaopt = \frac{B}{\eta \cdot D \cdot \tepochopt(\tpp)}
              = \frac{B \cdot \tpp^{-\mtepoch}}{\ctepoch \cdot \eta \cdot D}
\end{equation}

\subsection{Experimental details}\label{sec:lambda_experimental}

We use a GPT2-like LLM~\citep{radford2019gpt2}, with ALiBi
embeddings~\citep{press2022alibi} and SwiGLU~\citep{shazeer2020glu}.
%with the GPT-2 vocabulary (size 50257), and a context length of 2048
%tokens.
%
We train on SlimPajama~\citep{cerebras2023slimpajama} and always
evaluate over a held-out set of 1.1B tokens.
We use AdamW and $\mup$, with $\mup$ HPs derived from a smaller proxy
model, and a linear LR schedule, with a 10\% warmup followed by
decay-to-zero~\citep{bergsma2025straight}.
\cref{sec:experimental_details} has full experimental details.

\subsection{Results: $\tepoch$ and scaling $\lambda$}\label{sec:tema_results}

\begin{figure}[H]
    \centering
    \begin{minipage}{0.33\textwidth}
        \includegraphics[trim={0.3cm 0.4cm 0.3cm 0.3cm}, clip, width=\linewidth]{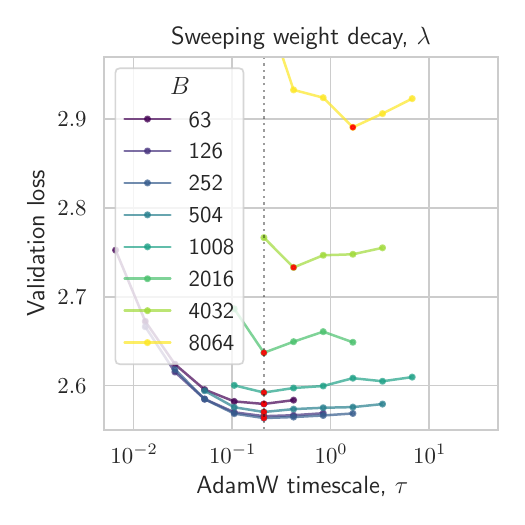}
    \end{minipage}\hfill
    \begin{minipage}{0.33\textwidth}
        \includegraphics[trim={0.3cm 0.4cm 0.3cm 0.3cm}, clip, width=\linewidth]{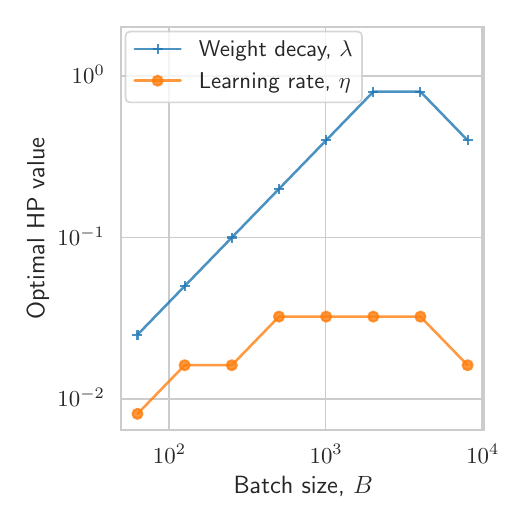}
    \end{minipage}\hfill
    \begin{minipage}{0.33\textwidth}
        \includegraphics[trim={0.3cm 0.4cm 0.3cm 0.3cm}, clip, width=\linewidth]{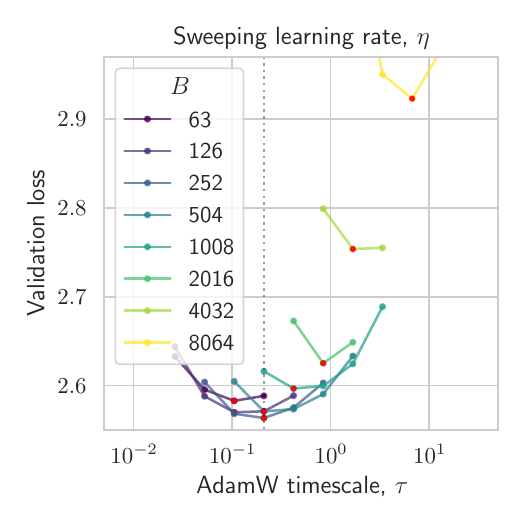}
    \end{minipage}
    \caption{(610M 20TPP): For each $B$, we sweep $\lambda$ and find
      $\tepochopt$ ($\leftfig$). $\tepochopt$ is stable around 0.21
      for $B \in [63, 2016]$, meaning $\lambdaopt$ scales linearly
      with $B$ over this range ($\middlefig$).  When sweeping $\eta$
      ($\rightfig$), the lower boundary over all curves is a bowl with
      a minimum at 0.21; the smallest $B$ settings have $\tepochopt$
      within 2$\times$ of this value, but as $B$ increases,
      $\tepochopt$ quickly drifts higher.\label{fig:tepochs}}
\end{figure}

\finding{The optimal $\tepoch$ remains stable as $B$ scales; $\lambdaopt$ scales linearly with $B$
(\cref{fig:tepochs}).}

As $B$ increases and $\lambda$ is tuned, we find that $\tepochopt$
remains roughly constant---i.e., changes in $B$ lead to commensurate
changes in $\lambdaopt$ (\cref{fig:tepochs}, $\middlefig$)---but only
up to a certain point, after which $\tepochopt$ begins to drift. The
drift point corresponds to the critical batch size $\bcrit$, above
which gradient information no longer scales linearly with $B$ and
diminishing returns set in (\cref{sec:bcrit}).%
\footnote{Training with $B > \cbs$ can be viewed as training with less
\emph{effective} data, thus decreasing the TPP, and shifting
$\temaopt$ higher (by our power law) and thus $\lambdaopt$
\emph{lower}.
A decrease in the optimal learning rate $\etaopt$ when $B > \cbs$ has
been previously observed~\citep{li2024surge,filatov2024time}.  These
works measure $\etaopt$ at very large batch sizes, e.g., $10\times$ or
$100\times$ $\bcrit$, which was not practical at the scales that we
trained.  Nevertheless, our observation of a similar ``surge''
phenomenon with $\lambdaopt$, occurring in AdamW rather than vanilla
Adam, motivates further study.}

If $\eta$ is tuned instead, $\etaopt$ fails to scale with $B$ up to
$\bcrit$ (\cref{fig:tepochs}, $\middlefig$); we observe instead a
certain maximum $\eta$, above which training becomes unstable; above
this, loss spikes occur from which training does not recover.
Consequently, $\eta$ has less flexibility to scale with $B$; training
stability is more fundamental than timescale.

In general, LLMs typically train faster and utilize hardware better
with larger $B$, but only up to $\bcrit$: beyond this point, much more
data (and compute) is needed to obtain the same loss, without
meaningfully reducing the total number of sequential training steps
(the poor trade-off for $B > \bcrit$ is depicted in
\cref{fig:cbs_panels}). Furthermore, \cref{sec:bcrit} will show that
there is also an \emph{optimal} batch size, $\bopt$, below which loss
is worse and utilization/parallelism suffer (because batches are
small). In practice, LLMs should therefore be trained in the regime
$\bopt \le B \le \bcrit$. Notably, this is precisely the range where
we have shown weight decay to scale predictably with batch size; our
findings therefore support the direct optimization of weight decay in
the most practically relevant training regimes.

\finding{With AdamW, we should adjust $\lambda$, not $\eta$, as $B$ changes.}

\begin{table}
  \centering
  \caption{(610M, 20TPP) Validation losses comparing tuning $\lambda$
    vs.\ $\eta$ across $B$ (data from \cref{fig:tepochs}).\label{tab:lr_v_wd}}
\begin{tabular}{@{}ccr@{}cccccccc@{}}
\toprule
$\hateta$                      & $\lambda$ & $B$$=$                   & 63         & 126        & 252        & 504        & 1008       & 2016       & 4032       & 8064       \\ \midrule
$\maxlr$                       & 0.1       &                    & 2.595          & 2.570          & \textbf{2.563} & 2.573          & 2.599          & 2.649          & 2.755          & 2.923          \\
Tuned                          & 0.1       &                    & 2.583          & 2.570          & \textbf{2.563} & 2.571          & 2.597          & \textbf{2.625} & 2.754          & 2.923          \\
$\maxlr$                       & Tuned     &                    & \textbf{2.579} & \textbf{2.565} & \textbf{2.563} & \textbf{2.570} & \textbf{2.592} & 2.637          & \textbf{2.733} & \textbf{2.891} \\ \bottomrule
\end{tabular}
\end{table}

Since tuning at scale is infeasible, we need a recipe for selecting
HPs in advance.  Unlike $\eta$, optimal $\lambda$ follows a
predictable relationship with $B$ (\cref{fig:tepochs}, $\middlefig$),
making $\lambda$ the more \emph{viable} target for real-world
adjustment.
Moreover, since $\eta$ has less flexibility to maintain optimal
timescale, we hypothesize adjusting $\lambda$ could also be
more \emph{effective}.
We tested this by comparing either tuning $\eta$ (using a default
$\lambda$=0.1---standard practice in LLM
pre-training~\citep{hoffmann2022empirical,brown2020language,almazrouei2023falcon,alephalpha2024introducing})---or
tuning $\lambda$ (using the $\mup$ proxy-tuned $\eta$).
Tuning $\lambda$ was strictly superior in 6 of 8 cases (\cref{tab:lr_v_wd}).
%
%This table also shows the drawbacks of using default HP settings;
%models obtain over 1\% worse error at large and small $B$---which
%drastically affects $\cbs$ estimation.
%
%For $B$=1008, we also swept the full cross-product of $\eta$
%vs.\ $\lambda$ values; we found tuning $\lambda$ alone remains
%optimal.

We also compared adjusting $\lambda$ versus $\eta$ as $D$ changes.
For a 111M 200TPP model, default HPs obtain a loss of 2.810, tuning
$\eta$ achieves 2.808, and tuning $\lambda$ obtains 2.805.  While
differences are small, the key point is that, when scaling $B$ or $D$:
optimizing $\lambda$ alone is viable and effective.

\finding{$\temaopt$ decreases as a power law in TPP; the law holds at scale (\cref{fig:hook}, $\leftfig$).}

%We now evaluate our approach to determining optimal $\lambda$ values
%via the optimal $\tepoch$.
At each $N$ and $D$, we calculated $\tepochopt$ over all $(\lambda,
B)$ pairs; we then fit \cref{eq:tepoch_scaling} to the results.
Full details are in \cref{sec:appendix_tepoch_fitting}.
A precise power law emerges ($\Rtwo$=0.975), with an optimal
$\tepoch$ around 1.0 at 1 TPP, decreasing to 0.01 at 1000 TPP\@.
10\textsuperscript{th} and 90\textsuperscript{th} percentiles of
fitted $\mtepoch$ over all points are (-0.529, -0.507) (computed as
in~\citet{hoffmann2022empirical} by bootstrapping: re-fitting on 80\%
of points, 1000$\times$), indicating a reliable power-law trend.
A decreasing $\tepoch$ stands in contrast to the prescription
of~\citet{wang2024how} (for multi-epoch training), who advocated
keeping $\tepoch$ constant as $D$ changes.

\cref{fig:hook} ($\leftfig$) includes four (labeled) points not used
in fitting, but computed later to evaluate predictive ability.
Even though some of these points are \emph{interpolated} in terms of
the fitting range (i.e., TPP range), they nevertheless represent much
greater scales; e.g., training a 3.3B-30TPP model requires
1000$\times$ the FLOPs compared to training the 111M-20TPP model
(whose data point is plotted nearby).
In other words, the law generalizes across at least 3 orders of
magnitude in compute.
%
%pEncouragingly, fitting 111M models alone would have provided a
%reasonable power law for predicting at these much higher scales.

%As noted in \cref{sec:cbs_experimental}, for training 1.7B 80TPP
%models, we leverage our $\tepoch$ estimates to set $\lambda$ directly
%without tuning.  We had previously trained with $\lambda$ set to the
%default value of 0.1 and obtained a loss of 2.184.  Using the
%$\tepoch$ estimate, we re-trained the model with the projected optimal
%$\lambda$ of 0.04, and obtained 2.179, a 0.2\% improvement.
% maybe only if asked?

\paragraph{Discussion}

The $\tepoch$ scaling law also predicts previously-observed HP scaling
in the literature.  E.g., for a fixed $N$,
\cref{eq:tepoch,eq:tepoch_scaling} together imply:
%\begin{equation}\label{eq:eta_scaling}
$\etaopt = B \cdot \cetad \cdot D^{\metad}$
%\end{equation}
, where $\cetad$ and $\metad$ are parameters.  This matches
Equation~(1) in~\citet{shen2024power}, and our implied $\metad$ is
close to their fit value (see \cref{sec:appendix_power_scheduler}).
\citet{bjorck2024scaling} also scaled $\eta$ as a power law in $D$.
They note that fitted power law exponents are similar when $B$ is
doubled, although the optimal $\eta$ is ``higher''.  More precisely,
we can see from their Figure~13 that the optimal $\eta$ appears to, in
fact, also double---consistent with the derived $\etaopt$ equation
above.

\takeaway{With AdamW, you can find $\tepochopt$ for a small $N,
  D$ by tuning $\lambda$.
  From there, $\tepochopt$ scales $\propto (\nicefrac{D}{N})^{-0.5}$.
  At larger $N$, $D$, set $\lambdaopt$ via
  \cref{eq:lambdaopt} and enjoy well-tuned models.}

\section{Scaling of optimal batch size $\bopt$ and critical batch size $\bcrit$}\label{sec:bcrit}
We now develop methodology to estimate $\bopt$ and $\cbs$ over the
dimensions of total training tokens $D$, total compute FLOPs $C$, and
validation loss $L$. Results show power-law scaling of both $\bopt$
and $\cbs$ with $D$, enabling estimation of $\cbs$ at scale via a
small number of test runs at modest budgets.

\subsection{Background: $\bopt$ and $\bcrit$}\label{sec:cbs_background}

\paragraph{$\bopt$}

As noted above, recent work has pursued an \emph{optimal} $B$: the $B$
achieving lowest loss given $N, D$.
\citet{hu2024minicpm} fit $\bopt$ using a power law in
(estimated) loss, and use $\mup$ to set $\eta$.
%
%$B$ and $\eta$ can be tuned together at small scales and transferred
%to larger runs~\citep{sengupta2023jais},
%
Joint power laws for optimal $\eta$ and $B$ have also been
fit~\citep{bi2024deepseek,porian2024resolving,carbonneaux2025cwm}.
E.g.,
\citet{bi2024deepseek} estimated \mbox{$\bopt = 0.292 C^{0.3271}$} (in tokens);
we refer to this fit as $\bdeepseek$ below.
\citet{li2025predictable} found $\etaopt$ to scale in $N, D$, while $\bopt$ primarily scales in $D$.
Qwen2.5~\citep{yang2024qwen2_5} also report studying how $\etaopt$ and
$\bopt$ scale with $N$ and $D$ (across dense and mixture-of-expert
LLMs), but without further details.

%jais: ''we tuned the optimal values for batch size and learning rate
%on a 40M-parameter model, and transferred the best values to our
%13B-parameter model.''

\paragraph{$\bcrit$}

Let $D$ be the number of training tokens required to reach target loss
$\lhat$ when using a batch size of $B$.  \mbox{$S = {D}/{B}$} is the
corresponding number of optimization steps.
Doubling a small $B$ doubles per-step gradient information; $\lhat$
can be reached in half the optimization steps, using the same total
$D$ (so-called ``perfect
scaling''~\citep{ma2018power,shallue2019measuring}).
But as $B$ increases further, step-wise gradient information becomes
more and more redundant: eventually, much larger $D$ is required to
reach $\lhat$, and $S$ decreases only marginally.
\citet{mccandlish2018empirical} show that $\langle D, S \rangle$ pairs
can be well fit by the equation:
\begin{equation}\label{eq:tradeoff}
S/\smin - 1 = (D/\dmin - 1)^{-1}
\end{equation}
where $\smin$ and $\dmin$ are parameters to be fit.  Intuitively,
$\dmin < D$ is the asymptotically minimum number of tokens that can
reach $\lhat$ (achieved with $\bopt$) and $\smin < S$ is the
asymptotically minimum number of \emph{steps} that can reach $\lhat$
(achieved as $B \rightarrow \infty$).
%These are also the asymptotes in \cref{fig:d_s_cartoon}.
\cref{eq:tradeoff} defines a hyperbolic curve, like those in
\cref{fig:cbs_panels}, where $B$ controls the position on the curve
and can be set depending on the importance of time (higher $B$
$\rightarrow$ higher $D$, lower $S$) or compute (lower $B$
$\rightarrow$ lower $D$, higher $S$).

\begin{definition}
\label[definition]{def:cbs}
The \emph{critical batch size} at $\lhat$ is defined from the fit
of \cref{eq:tradeoff} as $\cbs$=${\dmin}/{\smin}$.
\end{definition}

%; it is the largest \emph{useful} $B$ for
%training: the largest $B$ before $D$ rapidly increases.

From \cref{eq:tradeoff}, we can derive (\cref{sec:cbs_derivations}),
for a given $B$, the $D$ needed compared to $\dmin$:
\begin{equation}\label{eq:s_extra}
D = \dmin(1 + B/\cbs)
\end{equation}
\cref{eq:s_extra} implies that when $B$=$\cbs$, we require
$2\times\dmin$ tokens (and $2\times\smin$ steps) to reach $\lhat$.  $\cbs$ is a
\emph{transition} point along the $D$ vs.\ $S$ curve: for $B
> \cbs$, much higher $D$ is needed for only small reductions in $S$
(\cref{fig:cbs_panels}).
%
% McCandlish also refer to $\cbs$ as the largest \emph{useful} batch size.
%
%In this way, $\cbs$ is not ``optimal'' in the sense of achieving the
%lowest loss for a given $N$ and $D$.  Rather,
\citet{kaplan2020scaling} refer to $\bcrit$ as the \emph{optimal compromise}
between time and compute.  They determine $\cbs$ at smaller scales and
fit a power law for $\cbs$ as a function of $\lhat$.

\cref{eq:s_extra,eq:tradeoff} also imply $\bopt$ is theoretically equal to 1.
%, i.e., the batches should be more and more compute-efficient.
In practice, loss degrades below a particular
$\bopt$~\citep{hu2024minicpm,bi2024deepseek,porian2024resolving}, a
finding that ``appears to contradict the conventional wisdom'' about
$\bcrit$~\citep{porian2024resolving}.
%our results appear to contradict the conventional wisdom about the
%existence of a critical batch size [49, 61, 36] below which every batch size is good, finding instead% an
%optimal batch size below which performance degrades.
%
With well-tuned $\lambda$, we find small differences in loss across
small $B$, suggesting \cref{eq:tradeoff} may nevertheless provide a
good fit to observed data. \cref{sec:limitations} has further
discussion.

In recent work,
\citet{zhang2024how} define $\bcrit$ as the point where \mbox{$D = 1.2 \times \dmin$}.
They use a different training setup,
with a constant LR, weight averaging, and no weight decay.  Notably,
they observe little change in $\bcrit$ as $N$ varies at fixed $D$,
but, for a 302M model, find power-law scaling in $D$ as
\mbox{$\bcrit = 22.91 D^{0.47}$} (in tokens),
consistent with observed scaling across models at fixed TPP\@.
See \cref{sec:appendix_zhang} for further differences
with~\citet{zhang2024how}.

\subsection{Methods: estimating $\bopt$ and $\cbs$, and their scaling}\label{sec:cbs_methods}

\begin{figure}
  \noindent
\begin{minipage}[b]{0.32\textwidth}
  \centering
  \input{figures/fig_cbs_fitting.tex}
\end{minipage}%
\hfill
\begin{minipage}[b]{0.64\textwidth}
  \centering
  \input{figures/fig_cbs_panels.tex}
\end{minipage}
\end{figure}

\paragraph{Estimating $\bopt$}

We use the same experimental settings as \cref{sec:lambda_experimental}.
%
%As in \citet{shallue2019measuring}, ``We try to avoid making
%assumptions about how the optimal metaparameters vary as a function of
%batch size'' so even though we found good rules for this, we actually
%tuned HPs at each batch size at most of our model scales.  More
%details in appendix.
%
To ensure good HPs, we sweep $\lambda$ by factors of $2\times$ at each
$B$, $D$, $N$, except at the largest scales (see
Appendix \cref{tab:train_steps}) where we set $\lambda$ via the
projected value from \cref{eq:lambdaopt}.
In all figures, $B$ is reported in units of \emph{sequences}.

\paragraph{Estimating $\cbs$}\label{sec:estimating_bcrit}

%We first discuss how we fit $\cbs$ for a particular model with a
%particular loss target, $\lhat$.  We then describe how we fit a scaling
%law for $\cbs$.

Unlike $\bopt$, measuring $\cbs$ requires training models with
different $B$ \emph{to the same $\lhat$}.
Unfortunately, we do not know \emph{a priori} how many steps are
required to reach $\lhat$, yet we need this information to configure a
LR schedule that reaches its minimum value on the final step (the
typical setup, shown to be consequential in prior
work~\citep{hoffmann2022empirical,hagele2024scaling}).
Unfortunately, it is not feasible to \emph{search} for the precise
steps needed, i.e., by conducting training runs with different
schedules/step budgets.

\citet{mccandlish2018empirical} address this issue by performing
a single training run at a constant LR, while \citet{zhang2024how}
also use a constant LR, but use weight averaging to frequently
generate higher-quality checkpoints for evaluation
(\cref{sec:appendix_zhang}).

In contrast, we desired a method agnostic to the LR schedule.
We achieved this by fitting batch-size-specific power laws that model
how \emph{loss} scales with $D$.
These laws allow us to accurately \emph{interpolate} the $D$ required
to reach $\lhat$.  \cref{fig:cbs_fitting} depicts, for different $B$
and a given $\lhat$, the interpolated $D$ values (intersection points
of arrowed line and fitted loss curves).  The full process to obtain
$\cbs$ at $\lhat$ is:
\begin{enumerate}
\item For each $B$, train over different $D$, and
subsequently fit a $B$-specific power law \mbox{$L_{B}(D) = \irr + \dconst D^{-\beta}$}
on the resulting loss values (fitted curves in \cref{fig:cbs_fitting}).
\item Use fitted $L_{B}(D)$ to infer the $\dinfer$ needed to reach $\lhat$ as:
\mbox{$\dinfer = L_{B}^{-1}(\hat{L}) = (\nicefrac{\dconst}{\hat{L}-\irr})^{\frac{1}{\beta}}$}.
\item Fit \cref{eq:tradeoff} on the resulting $\langle \dinfer, S$=${D}/{B} \rangle$ pairs, and obtain $\cbs = {\dmin}/{\smin}$.
\end{enumerate}
This method makes no assumptions about the LR schedule or optimizer,
while enabling measurement of $\cbs$ at arbitrary losses without
re-training.
\cref{sec:appendix_estimating_cbs} provides further details,
including fits of $L_{B}(D)$ at other model scales
(\cref{fig:quad_cbs_fitting}) and a summary of the full procedure
(\cref{alg:cbs_fitting}).

\paragraph{$\bopt$ and $\cbs$ scaling}

We collect $\bopt$ across different $N$ and $D$, and fit a power law
in both data $D$ and compute $C$ (via the standard approximation
$C \approx 6ND$~\citep{kaplan2020scaling,hoffmann2022empirical}).
For $\bcrit$, we use the procedure described above to estimate $\cbs$
across multiple $\lhat$, across different $N$.  From each $\cbs$
estimate, we obtain a pair $\langle \dmin, \cbs \rangle$.
We propose that $\cbs$ follows a power law in $\dmin$, according to:
\begin{equation}\label{eq:cbs_scaling}
  \cbs(\dmin) = {\ccbs} \cdot \dmin^{\mcbs}
\end{equation}
Where $\ccbs$ and $\mcbs$ are fit on the $\langle \dmin, \cbs \rangle$
pairs.
%In the following results, we find strong fits
%to \cref{eq:cbs_scaling} over experimental data.

\subsection{Results: $\bopt$ and $\bcrit$}\label{sec:results_cbs}

\begin{figure}[H]
    \centering
    \begin{minipage}{0.33\textwidth}
        \includegraphics[trim={0.3cm 0.4cm 0.3cm 0.3cm}, clip, width=\linewidth]{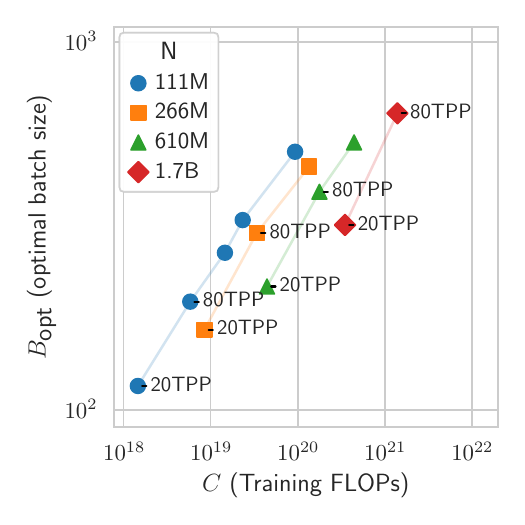}
    \end{minipage}\hfill
    \begin{minipage}{0.33\textwidth}
        \includegraphics[trim={0.3cm 0.4cm 0.3cm 0.3cm}, clip, width=\linewidth]{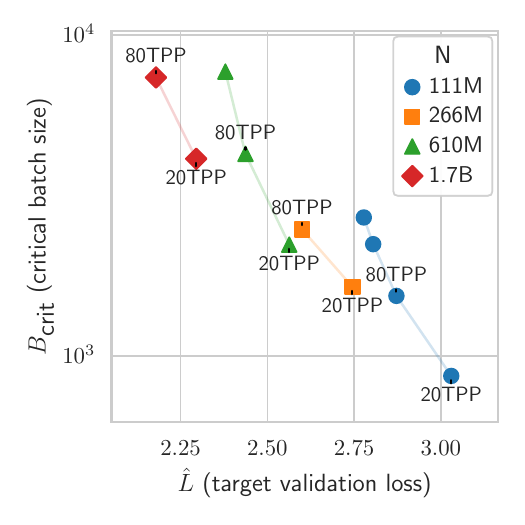}
    \end{minipage}\hfill
    \begin{minipage}{0.33\textwidth}
        \includegraphics[trim={0.3cm 0.4cm 0.3cm 0.3cm}, clip, width=\linewidth]{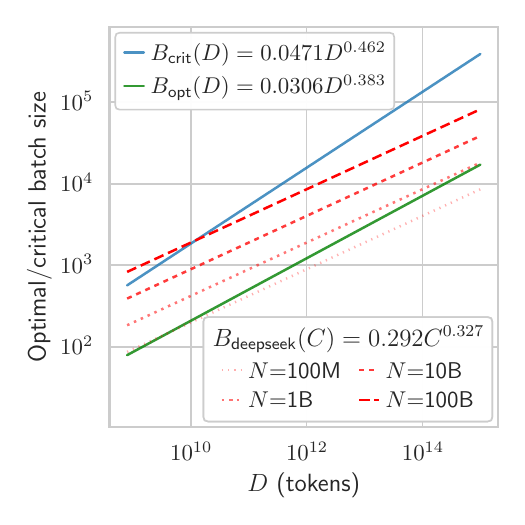}
    \end{minipage}
    \caption{Prior work suggests $\bopt$ scales in $C$ ($\leftfig$)
      and $\bcrit$ in loss ($\middlefig$), but \emph{this only holds
      at a fixed $N$/TPP} (same data as \cref{fig:hook}
      $\middlefig$/$\rightfig$); \cref{fig:hook} shows scaling in $D$
      is the fundamental relationship.
      Plotting $\bdeepseek(C)$ law from~\citet{bi2024deepseek}
      ($\rightfig$), but over $D$ (using $C \approx 6ND$ to obtain $C$
      for the spurious dependence on $N$), we see
      \citet{bi2024deepseek} used generally efficient $B$ values
      (i.e., within the $\bopt < B < \bcrit$ regime) despite fitting $C$
      rather than $D$ ($\bopt$ and $\bcrit$ lines from
      \cref{fig:hook}).\label{fig:related}}
\end{figure}

\finding{\cref{eq:tradeoff} provides a decent fit to the trade-off
  between training time and compute.}

Across different model scales and loss targets, we consistently find
that our $\langle D, S \rangle$ pairs fit \cref{eq:tradeoff} well
(examples in \cref{fig:cbs_panels}, appendix
\cref{fig:quad_cbs_fits}).
Fits are worse at very small $B$, as noted above: smaller batches are
not monotonically more efficient; \cref{sec:limitations} discusses
some potential reasons for this.

% Notably, the horizontal alignment of similar $\cbs$ in
%\cref{fig:cbs_panels} already suggests $\cbs$ does not depend on the
%loss (in color), but rather on number of tokens processed (shared
%y-axis).

%\input{figures/fig_cbs_scaling_zhang.tex}

\finding{$\bopt$ and $\cbs$ obey power laws in $D$ and $\dmin$, not in $C$ or $L$.}

$\bopt$ and $\cbs$ datapoints fit power laws quite well
(\cref{fig:hook}, $\middlefig$, $\Rtwo$=0.984) and (\cref{fig:hook},
$\rightfig$, $\Rtwo$=0.940).
10\textsuperscript{th} and 90\textsuperscript{th} percentiles over all
points are (0.367, 0.391) for fitted $\mbopt$ and (0.491, 0.526) for
$\mcbs$ (computed as in \cref{sec:tema_results}).  Note $\mcbs$ is
higher when fitted over \emph{all} points (as opposed to only
small-scale runs), partly reflecting the 111M points trending lower as
TPP increases.

Our fitted $\bcrit$ power law exponent is very close to that
from~\citet{zhang2024how}: 0.47 vs.\ 0.462.  Given the many
differences in approach (including dataset, use of weight decay, LR
schedule, etc., \cref{sec:appendix_zhang}), this agreement suggests
the fundamental relationship of $\bcrit$ with $D$ persists across such
differences.
%Moreover, given the significant implications of $\bcrit$'s
%fundamental scaling behavior, it is valuable, scientifically and
%practically, that different approaches independently measure a
%similar scaling relationship.

\cref{fig:related} ($\leftfig$) plots $\bopt$ versus $C$ and
\cref{fig:related} ($\middlefig$) gives $\bcrit$ versus $L$ using
the same data as in \cref{fig:hook}.
In each case, a power law does not fit all points (as proposed
previously), but points at the same $N$, or same TPP, can roughly be
linked by (parallel) lines.  This is a consequence of power-law
scaling in $D$ (see \cref{sec:appendix_loss_scaling}).
That is, scaling in $D$ is the fundamental scaling relationship:
$\bopt$ and $\bcrit$ both scale in $D$ regardless of TPP, model size,
or loss---it is only when using another (misleading) scaling factor
such as $C$ or $\lhat$ that TPP or model size appears important, as in
these plots.

\cref{fig:related} ($\rightfig$) compares the recommended batch sizes
from $\bdeepseek$ to those from $\bopt$ and $\bcrit$.  Since
$\bdeepseek$ scales in $C$, it is larger for larger $N$.  Over a range
of modern model sizes, $\bdeepseek$ values generally fall between our
projected $\bopt$ and $\bcrit$, varying in the extent to which they
are compute-efficient (close to $\bopt$) or time-efficient (close to
$\bcrit$).
%In the next section, we develop a targeted approach to find $N$, $D$,
%and $B$ settings achieving an optimal balance of time and compute.

\finding{Weight decay affects the accuracy of fitted batch size scaling laws.}\label{finding:wd_bopt}

Prior work has typically held $\lambda$ fixed when fitting batch-size
scaling laws \citep{bi2024deepseek,porian2024resolving,zhang2024how}.
Doing so not only degrades loss (\cref{sec:lambda}) but also reduces
the accuracy and generality of the fitted scaling relationships.  We
demonstrate this in appendix \cref{tab:wd_bopt}: rather than tuning
$\lambda$ for each $B$, we train with several fixed $\lambda$ values
across all runs.
As \cref{tab:wd_bopt} shows, increasing $\lambda$ systematically
raises the estimated $\bopt$.  This arises because the fundamental
scaling variable is the AdamW timescale $\tema = B /
(\eta \lambda D)$: when $\lambda$ increases, the batch size that
minimizes loss must increase proportionally to preserve the optimal
$\tema$.
In Appendix~\ref{sec:app_wd_affects_b}, we show that these effects
distort the fitted power-law slope and reduce fit quality ($R^{2}$),
leading to scaling laws that do not generalize to large-scale
training---even if the same fixed weight decay is used there.  When
$\lambda$ is tuned to maintain the optimal timescale (final row
of \cref{tab:wd_bopt}), the resulting $\bopt$ follows a clean and
accurate power law.

Similar distortions occur for $\bcrit$ when $\lambda$ is fixed 
(Appendix~\ref{sec:app_wd_affects_b}).

\takeaway{You can estimate $\bopt$ and $\cbs$ for a small $N$ by training with
  different $B$, $D$ and $\lambdaopt$, and computing loss.  From
  there, $\bopt \propto D^{0.4}$ and $\cbs \propto D^{0.5}$.  At
  larger $N$ and $D$, \cref{eq:s_extra} lets you estimate trade-offs in
  FLOPs ($\propto D$) vs.\ training time ($\propto S = {D}/{B}$) at
  different $B$.}

\section{Training settings for balancing time and compute}\label{sec:balancing}
\subsection{Background: compute-optimal and overtrained models}\label{sec:balancing_background}

Given a fixed training FLOPs budget, $C$, how should we allocate model
size $N$ versus number of training tokens $D$ in order to minimize
\emph{loss}?
\citet{hoffmann2022empirical} propose to model loss as:
\begin{equation}\label{eq:chinchilla}
L(N,D) = E + \nconst N^{-\alpha} + \dconst D^{-\beta}
\end{equation}
$\nconst$, $\alpha$, $\dconst$, and $\beta$ are parameters fit on
observed training runs.  From \cref{eq:chinchilla},
\citep{hoffmann2022empirical} derives functions for loss-optimal
$\nopt(C)$ and $\dopt(C)$ (constraining $L(N,D)$ by $C \approx 6 N
D$).
Results indicate $\nopt$ and $\dopt$ scale roughly equally as $C$
increases, with the optimal $\nicefrac{D}{N}$ ratio relatively
constant at around 20 TPP\@.  Replication studies have found similar
results~\citep{besiroglu2024chinchilla,porian2024resolving},
and 20 TPP has become a rule-of-thumb for compute-optimal
training~\citep{dey2023cerebras,zhang2024how}.
%
%In this work, we call models \emph{undertrained} (or
%\emph{overtrained}) when trained below (or above) 20 TPP\@.

Overtrained, inference-efficient
models~\citep{touvron2023llama,biderman2023pythia,dubey2024llama} have
largely trained with similar batch sizes to those used in
compute-optimal training; such efforts should now consider training
with much greater data parallelism, leveraging our finding that
$\bopt$ and $\bcrit$ will be higher given the higher training $D$.

%Discuss power laws without irreducible loss, e.g. in \citet{kaplan2020scaling}
%
%\begin{equation}\label{eq:powerlaw}
%L(x) = c x^{m}
%\end{equation}

\subsection{Methods: exploring the trade-offs of FLOPs vs.\ time}

%We can use our $\cbs$ power law to help explore the tradeoffs of FLOPs
%vs.\ time in LLM pre-training.
%
%However,
To compare models of different sizes on a common temporal
axis, we must map number-of-optimization-steps to a common temporal scale.
Our initial approximation is
%\begin{equation}\label{eq:ttime}
\mbox{$\ttime \propto \nicefrac{\totflopstext}{B}$},
%\end{equation}
which is also FLOPs \emph{per token} times number of steps.  E.g., if
\mbox{$\flops \approx 6ND$}, \mbox{$\ttime \approx {6ND}/{B} = 6N \cdot S$}.  This
aligns well with our measured runtimes: doubling N doubles step time;
doubling B halves wall-clock time (for the same $S$).
%
%Note $\nicefrac{\flops}{B}$
%correlates well with real measurements of our training runs: doubling
%$N$ (and thus $\flops$) makes each step take roughly twice as long,
%while doubling $B$ cuts wall-clock training time in half.}
%Indeed, ``data parallelism can achieve near-perfect scaling at
%small scales''~\citep{smith2022using}.}

Now, assume a model of size $N$ can train to loss $\lhat$ using
$\dmin$ tokens (here \emph{min} denotes using $\bopt$).
Let us refer to $N$ and $\dmin$ as a \emph{base setting}.
A variety of $N$, $\dmin$ pairs can reach $\lhat$ in the $\bopt$
setting, from small models trained on many tokens, to large models
trained on fewer tokens.  \citet{hoffmann2022empirical}~refers to
these as iso-loss \emph{contours} of \cref{eq:chinchilla}.
Suppose a given base setting requires $\minflops$ FLOPs.
From this setting, we may increase $B$ to decrease training time
(fewer steps), but \cref{eq:s_extra} indicates a need for $(1 +
{B}/{\cbs})$ extra \emph{data} in order to reach the same $\lhat$.  If
FLOPs is linear in $D$ (as in $C = 6ND$), we will require the same
proportion of extra FLOPs, i.e.,
\begin{equation}\label{eq:flops_extra}
  \totflops = \minflops(1 + {B}/{\cbs(\dmin)})
\end{equation}
where $\totflops$ denotes the total FLOPs needed at $B > \bopt$, and
$\cbs(\dmin)$ captures that the excess FLOPs depends on $\bcrit$,
which itself scales with $\dmin$.  In other words, the base setting
dictates $\bcrit$, and $B/\bcrit$ dictates the excess FLOPs.
%Intuitively, training to $\lhat$ with $B > \bopt$ saves time but costs
%FLOPs.

Consider a target FLOP budget of $\totflops = \hatf$ and the goal of
reaching $\lhat$ \emph{as fast as possible}.
Since time $\propto \nicefrac{\totflopstext}{B}$, time is minimized by
maximizing $B$.  However, by construction, $B$ is not a free variable:
it is constrained by \cref{eq:flops_extra} and can be expressed as a
function of $N$ and $\dmin$:
\begin{equation}\label{eq:b_equation}
  B(N,\dmin) = \left(\frac{\hatf}{\minflops} - 1 \right)\cbs(\dmin)
\end{equation}

Time is therefore minimized by finding $N$, $\dmin$ that maximize this
function (over all the $N$, $\dmin$ that train to loss $\lhat$).
The ${\hatf}/C$ ratio is a measure of the \emph{excess} FLOPs that can
be spent toward increasing $B$; it is largest when $\minflops$ is
smallest, i.e., when $N$ and $\dmin$ is most compute-efficient (i.e.,
$N/\dmin \approx$ 20~TPP).
But \cref{eq:b_equation} as a whole captures an elegant tension
between compute efficiency and $\cbs$: we can maximize $B$ (and
minimize training time) by either
\begin{inparaenum}[(1)]
 \item minimizing the FLOPs of the base setting (generating more
   excess FLOPs for increasing $B$), or
 \item maximizing $\dmin$ (overtraining, which increases $\cbs(\dmin)$).
\end{inparaenum}
For a given $\hatf$, either (1) or (2) may take
precedence.
%, depending
%on how compute efficiency and $\cbs$ scale relative to $\dmin$.

We use the following procedure to explore the time vs\@. compute
Pareto frontier for a target loss $\lhat$:
% Our full process for generating FLOPs vs.\ time curves is:
\begin{enumerate}
  \item Fit \cref{eq:chinchilla} on our $\bopt$ training runs.
    Express resulting $L(N,\dmin)$ as ${\dmin}_{\lhat}(N)$.
  \item Using ${\dmin}_{\lhat}(N)$, get contour points $\langle N,
    {\dmin} \rangle$ of the given $\lhat$.
    %
    % D = \left(\frac{\dconst}{\lhat - E - \nconst N^{-\alpha}}\right)^{\frac{1}{\beta}}
    %
    Each such pair consumes $\minflops \approx 6 N \dmin$
    FLOPs and takes $\nicefrac{\minflops}{B}$ time (rightmost points
    on \cref{fig:paretos} curves).
  \item Use \cref{eq:flops_extra} to compute $\totflops$ as we scale
    $B$ (crucially, using the estimate of $\cbs$ from fitted
    \cref{eq:cbs_scaling}), and generate further points along each
    curve.
  \item The non-dominated points over all curves provide the time
    vs.\ compute Pareto frontier.
\end{enumerate}

\subsection{Results: balancing time and compute}

We carry out this procedure using model sizes of 150M, 210M, 550M,
1.1B, and 2.1B, and a loss target of $\lhat$=2.6, yielding iso-loss
contour points from 150M 600TPP to 2.1B 2TPP.  Our fit of
\cref{eq:chinchilla} yielded $\alpha$=$0.313$ $\approx$
$\beta$=$0.282$, giving an optimal TPP ratio of $\approx$20.6 at
$\lhat$=2.6.

\finding{Overtrained, but not undertrained, models are on the FLOPs
  vs.\ time Pareto frontier.}

\begin{figure}[H]
    \centering
    \begin{minipage}{0.33\textwidth}
        \includegraphics[trim={0.3cm 0.4cm 0.3cm 0.3cm}, clip, width=\linewidth]{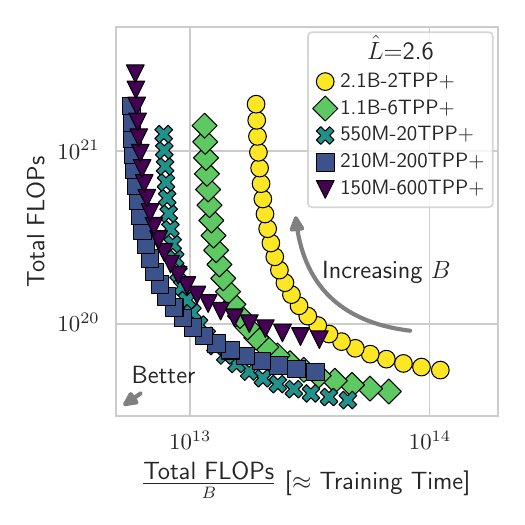}
    \end{minipage}\hfill
    \begin{minipage}{0.33\textwidth}
        \includegraphics[trim={0.3cm 0.4cm 0.3cm 0.3cm}, clip, width=\linewidth]{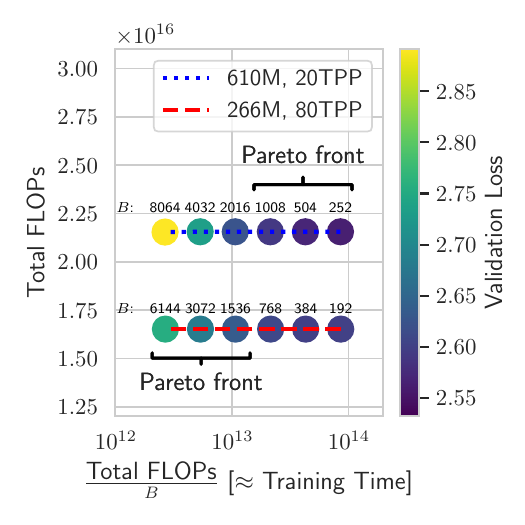}
    \end{minipage}\hfill
    \begin{minipage}{0.33\textwidth}
        \includegraphics[trim={0.3cm 0.4cm 0.3cm 0.3cm}, clip, width=\linewidth]{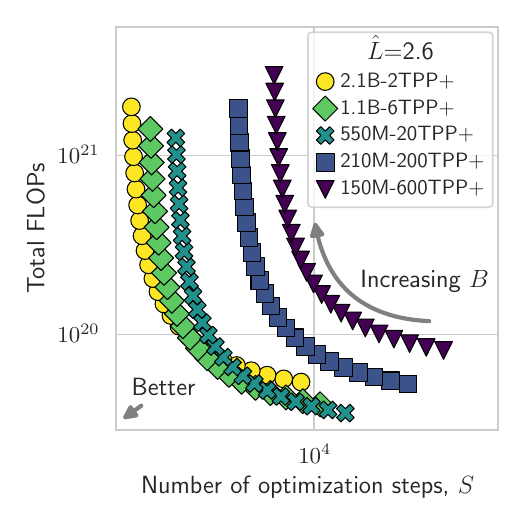}
    \end{minipage}
    \caption{($\leftfig$): Iso-loss curves illustrating time--compute
      Pareto frontier ($\lhat$=2.6). As $B$ increases along curves,
      more compute (y-axis), but less time (x-axis) is required. Here
      time $\propto \nicefrac{\totflopstext}{B}$.  ($\middlefig$):
      Observed runs where some overtrained models (red line) are on
      frontier: $L$ in color, $B$ labeled.  ($\rightfig$): Iso-loss
      curves, but where time = steps; a very different frontier
      emerges.\label{fig:paretos}}
\end{figure}

Specifically, when using $\ttime \propto \nicefrac{\totflopstext}{B}$,
we find overtrained models are FLOP-optimal at certain time budgets
(\cref{fig:paretos}, $\leftfig$).  Compute-efficient 20~TPP are
optimal in pure FLOPs (i.e., ignoring time), as expected, while
compute-efficient and overtrained models dominate undertrained ($<$ 20
TPP) models in time and FLOPs.  Indeed, this is expected from
\cref{eq:b_equation}: \emph{under}training reduces both the excess
FLOPs and $\cbs$ terms, and thus is never optimal with this model of
time.
%
%Additional interesting Pareto frontiers are given in
%\cref{sec:pareto_appendix}, including for other fits to
%\cref{eq:chinchilla} (from the literature), other loss targets, and
%other models of training time that account for both data and model
%parallelism.
% Refer to \cref{fig:pareto_steps_only} in
% \cref{sec:pareto_steps_only} for what happens if we consider
% all the FLOPs to be parallelizable, and why that isn't a good idea.

\finding{When using $B \gg \bopt$, it is Pareto-inefficient to train to 20~TPP.\label{finding:largeb}}

%We now discuss a somewhat subtle matter that has a very clear and
%important conclusion.

Notice that \cref{fig:paretos} adds ``\emph{$\ldots$TPP+}'' to curve
labels.  Here the $+$ sign is a reminder that as we increase $B$, we
require $(1 + {B}/{\cbs})$ \emph{extra data} to reach the
same $\lhat$; i.e., points with higher $B$ are trained to a higher
\emph{actual TPP} than the base setting.
%at $\tpp$=$\nicefrac{\dmin}{N}$.
For example, once the 2TPP+ curve in \cref{fig:paretos} reaches
10$\times$ its minimum FLOPs, it is \emph{actually} training at 20
TPP\@.
Since starting from an undertrained base setting is never Pareto
optimal (as just discussed above), \emph{it is \textbf{always}
suboptimal to train a model with a large $B$ to 20 actual TPP}.  If
large-batch training is needed, the configuration should start from a
20 TPP+ base setting and scale $B$ from there (to $>$20 TPP).

We can see this finding play out in real training runs.
\cref{fig:paretos} (middle) demonstrates observed runs where our 266M
80TPP models \emph{dominate} our 610M 20TPP models (i.e., in FLOPs
\emph{and} time)---when both train with large $B$.  (Note in this
plot, results are not iso-loss: the frontier is over $L$, $C$, and
time.)

\finding{The Pareto-optimal settings depend on the formulation of
  time/parallelism strategy.}

While $\nicefrac{\text{FLOPs}}{B}$ is a good model of data parallel
training, it does not incorporate the potential for model
parallelism~\citep{smith2022using}.
In the extreme we could assume that all $6N$ FLOPs could be executed
concurrently per input token.  Under this formulation, training time
is proportional only to the number of steps, regardless of model
scale.
\cref{fig:paretos} ($\rightfig$) shows the Pareto frontier that would
result from this formulation; if we pay no time cost for larger
models, we can train faster by using undertrained large models,
although, exactly as with overtrained models, they suffer in FLOPs.

While LLMs cannot be fully parallelized due to the inherent sequential
nature of a Transformer's layer-by-layer computation, this formulation
could be refined by incorporating depth or other architectural
features.
For example, \citet{inbar2024time} predict training time via linear
regression over total FLOPs and memory-copy operations, fit to real
(single-TPU) runs.
As formulations improve, different Pareto-optimal configurations will emerge.

\finding{Inaccurate $\bcrit$ scaling leads to inaccurate
  Pareto-optimal configurations.}

Because the Pareto frontier in \cref{sec:balancing} depends directly
on the $\bcrit$ power-law fit, any error in that fit produces
corresponding errors in the predicted trade-offs between training time
and compute.  Accurate $\bcrit$ estimation, in turn, depends on
effective $\lambda$ tuning
(\cref{finding:wd_bopt,sec:app_wd_affects_b}).  When $\lambda$ is
fixed as in standard practice (and thus the $\bcrit(D)$ slope is
misestimated), $\bcrit$ will be systematically over- or underpredicted
at scale, altering the computed frontier and the apparent
Pareto-optimal configurations.
To illustrate, artificially varying the $\bcrit$ exponent changes
which models appear on the frontier: as the exponent increases (and
$\bcrit$ rises), higher-TPP models move to the frontier; when $\bcrit$
is underestimated, only low-TPP (e.g., 20~TPP) models appear
Pareto-optimal.  Hence, an inaccurate $\bcrit$ scaling law produces
misleading frontiers and can lead to unexpectedly-longer training
durations and suboptimal compute allocations.

Recent work suggests that other estimators of $\bcrit$ (such as those
based on the gradient noise scale~\citep{mccandlish2018empirical}) can
also be systematically biased (\cref{sec:app_bcrit}), leading to
similar distortions in the Pareto frontier.

\takeaway{To balance time and compute at a target loss, select $(N,
  D_{\min})$ from \cref{eq:chinchilla}, determine $\bcrit(D_{\min})$
  via \cref{eq:cbs_scaling}, and use \cref{eq:flops_extra} to estimate
  compute for any $B$.  Under $\text{Time}\!\propto\!\text{FLOPs}/B$,
  the resulting time--compute trade-off favors higher $D$
  (overtraining) (\cref{fig:paretos}).}

\section{Conclusion}\label{sec:conclusion}

We have presented a comprehensive empirical study of hyperparameter
scaling laws in LLM pre-training, focusing on weight decay and batch
size. Our approach leverages the AdamW timescale ($\tepoch$) to
develop robust scaling relationships that predict optimal
hyperparameter settings across a broad spectrum of model ($N$),
dataset ($D$), and batch sizes ($B$). We demonstrated that optimal
$\tepoch$ decreases as a power law with the tokens-per-parameter
ratio, providing a systematic method to set weight decay optimally
across diverse training scenarios.

Furthermore, we introduced a novel, practical methodology for
estimating critical batch size ($\bcrit$). Our findings diverge from
influential prior work that tied $\bcrit$ predominantly to compute or
loss, while agreeing with the recent findings of~\citet{zhang2024how}
that underscore dataset size as the principal scaling factor.
Additionally, we showed that contrary to previous studies suggesting
optimal batch size ($\bopt$) scales primarily with compute, it also
exhibits a clear power-law dependence on $D$.

Also, our analysis of Pareto-optimal configurations reveals an
important strategic advantage for smaller, overtrained models in
scenarios where rapid training and high parallelism are prioritized.

\cref{sec:limitations} notes limitations and directions for
further study suggested by our results.
In particular, as inference-time scaling comes to the fore, inference
time and compute must also be considered as first-class Pareto
objectives.  Moreover, finer-grained configuration decisions, such as
model depth and context length, should be considered along with $N$,
$D$, and $B$.

\begin{ack}
  We thank the NeurIPS reviewers for their helpful feedback.  None of
  the authors received third-party funding or third-party support for
  this work.  None of the authors have financial relationships with
  outside parties that could potentially be perceived to influence
  this research.
\end{ack}

\bibliography{batches}
\bibliographystyle{cereb}

%%%%%%%%%%%%%%%%%%%%%%%%%%%%%%%%%%%%%%%%%%%%%%%%%%%%%%%%%%%%%%%%%%%%%%%%%%%%%%%
%%%%%%%%%%%%%%%%%%%%%%%%%%%%%%%%%%%%%%%%%%%%%%%%%%%%%%%%%%%%%%%%%%%%%%%%%%%%%%%
% APPENDIX
%%%%%%%%%%%%%%%%%%%%%%%%%%%%%%%%%%%%%%%%%%%%%%%%%%%%%%%%%%%%%%%%%%%%%%%%%%%%%%%
%%%%%%%%%%%%%%%%%%%%%%%%%%%%%%%%%%%%%%%%%%%%%%%%%%%%%%%%%%%%%%%%%%%%%%%%%%%%%%%
\newpage
\appendix

\section{Broader impacts}\label{sec:broader}
This paper presents methods to train LLMs more efficiently:
practitioners can use our methods to reduce the total compute FLOPs
used to train models, subject to time constraints.
Given the intense pressure to advance LLM capabilities as quickly as
possible, our methods can therefore reduce the associated
environmental and financial costs of LLM
training~\citep{patterson2021carbon,bender2021dangers}.

Moreover, hyperparameter tuning is a key contributor to these costs,
and impairs equity in AI research, as tuning success depends directly
on researcher finances~\citep{strubell2019energy}.
% For models that will be re-trained, \emph{sensitivity} to
% hyperparameters also leads to downstream
% costs~\citep{strubell2019energy}.
We hope our exploration of optimal hyperparameter scaling can reduce
the burden of hyperparameter tuning at scale and thus improve equity
in AI\@.

\section{Limitations}\label{sec:limitations}
While our findings corroborate prior work and provide strong evidence
for the proposed scaling laws in $\tepoch$, $\bopt$, and $\bcrit$,
there are several limitations that merit further study.

\paragraph{EMA perspective}

As the EMA perspective regards parameters $y_t$ as a function of
updates $x_t$, it fails to account for $x_t$ actually depending on
earlier values of $y_t$ (e.g., $y_{t-1}$). Yet although this
perspective has formal limitations, we nevertheless find it a
useful \emph{conceptual} model of training, as it predicts behavior
that is supported by experiments.

\paragraph{Optimization and training setup}

Our work focuses on AdamW (the standard optimizer for LLM training).
While the EMA perspective applies directly to other optimizers that
use decoupled weight decay, such as Sophia~\citep{liu2023sophia} and
MuonClip~\citep{kimi2025k2}, it may not apply to approximate second
order methods, e.g., Shampoo~\citep{gupta2018shampoo}.  However, it
can be used when applying AdamW (and related optimizers) in Shampoo's
eigenbasis, which was shown to be effective in
SOAP~\citep{vyas2024soap}.

We present results with a single (standard) learning rate schedule.
Our method for obtaining $\cbs$ estimates would be quite efficient
with a warmup-stable-decay (WSD)
schedule~\citep{hu2024minicpm,wen2024understanding}, as we could
perform a single training run with each batch size, but decay at
various milestones in order to get points along the scaling law,
essentially following the approach in~\citet{hagele2024scaling}, but
with separate laws for each batch size.

We used the maximal update parameterization in all experiments, which
generates a learning rate $\eta$ adjustment for each model width.  Our
results suggest this approach enables good models at arbitrary $N$,
$D$, and $B$ when combined with adjustments to $\lambda$.  This
strategy is informed by our experiments comparing re-adjusting $\eta$
vs.\ $\lambda$ in \cref{sec:tema_results}. However, it is not
feasible, at this scale, to verify whether substantially better models
could be obtained by sweeping the full cross-product of $\eta$ and
$\lambda$ values.

Our study specifically focuses on the practically important setting of
single-epoch LLM pre-training.  \citet{wang2024how} indeed noted
differences in optimal $\tema$ when using multi-epoch training,
possibly due to data repetition. Reconciling these differences by
isolating the effects of repetition versus scale is an interesting
follow-up direction.

Here we only experimented with a single dataset, vocabulary, and
context length.  We obtained a similar $\cbs$ scaling law
to \citet{zhang2024how}, but it would be interesting to see if
differences in the coefficient of our power laws could be attributed
to specific differences in approach (e.g., differences in dataset,
context length, learning rate schedule, use of weight decay,
etc.).  \cref{sec:appendix_zhang} has further discussion of
differences with \citet{zhang2024how}.

We have also not explored how changes in numerical precision could
affect scaling laws.  Recent work~\citep{kumar2024scaling} showed
that, in terms of scaling laws, lower precision reduces the model's
\emph{effective} parameter count.  This suggests precision would
have no impact on scaling of $\bopt$ or $\bcrit$, which do not scale
in $N$.  Lower precision, however, could increase the \emph{effective}
TPP (via smaller effective $N$), thereby altering $\temaopt$.

\paragraph{Small batches, large batches, and dynamic batch sizing}

We consistently find that smaller and smaller batches do not grow
asymptotically closer to $\dmin$, as predicted by theory, but
eventually degrade in loss.  One possibility is that $\lambda$ tuning
is not sufficient with very small $B$, and further tuning of other
hyperparameters may be needed, such as the Adam $\beta$ parameters (as
suggested in recent
work~\citep{porian2024resolving,zhang2024how,marek2025small}).
Some preliminary tests using the $\beta_2$ scaling rule
from \citet{marek2025small} showed loss improvements at small $B$.
Since we are unlikely to train with small batches at scale, and using
them even with smaller LLMs significantly impairs our ability to train
both efficiently and quickly, it is unfortunately difficult to justify
further exploration in this direction.

Regarding large batches, our methods do not account for the many
practical systems-related issues, including bandwidth and
communication overheads, memory limits of hardware, synchronization
delays, etc.  
Moreover, as batch sizes increase, techniques such as optimizer
sharding may be needed, which further complicate performance
model~\citep{almazrouei2023falcon}.
Our scaling laws do, however, explicitly define a practically relevant
regime of training batch sizes: $\bopt \le
B \le \bcrit$. Practitioners can leverage this identified regime
alongside system-specific profiling (e.g., evaluating utilization at
various batch sizes) to select optimal settings balancing algorithmic
and systems constraints.

Exploring optimal dynamic batch sizing is a natural future direction
for our work.  While the potential gains were found to be small in
theory by \citet{mccandlish2018empirical}, more recent work has found
significant wall clock speedups~\citep{meterez2025seesaw}.
%
%We also plan to explore dynamic measurement of the gradient noise
%scale~\citep{mccandlish2018empirical}, which has been shown to be
%correlated with the empirical $\cbs$, and may thus help guide a
%dynamic batch size schedule.

%- We also only consider training TIME and training COMPUTE, rather
%than inference TIME and inference COMPUTE.
%
%- overtrained models do take more training compute but reduce
%inference time and compute~\citep{touvron2023llama}, but we note we
%provide nuance to this idea by noting smaller can also reduce training
%TIME both because they use less time per optimization steps, and
%because they allow greater parallelism through a larger batch size.
%
%- Ideally, if we had an idea of our total expected training compute +
%inference compute needs, we could add inference compute to the y-axis
%on our Pareto plots and compare against training time, perhaps also
%incorporating any constraints we have on inference time.

\section{Experimental Details}\label{sec:experimental_details}

\begin{table}
  \centering
  \caption{Model architectures used in experiments\label{tab:model_info}}
\begin{tabular}{@{}ccccc@{}}
\toprule
Model & $d_{model}$ & $n_{layers}$ & $d_{head}$ \\ \midrule
111M  & 768    & 10      & 64           \\
266M  & 768    & 32      & 64           \\
610M  & 2048   & 10      & 64           \\
1.7B  & 2048   & 32      & 64           \\
3.3B  & 2048   & 64      & 64           \\ \bottomrule
\end{tabular}
\end{table}

\begin{table}
  \centering
  \caption{Models, tokens-per-parameter (TPP) and corresponding
    dataset sizes (in tokens) used in main experiments.  We also list
    the total number of batch sizes, $B$, trained at each scale and
    TPP, as well as the number of $B$ for which we tuned $\lambda$.
    For $B$ where $\lambda$ was not tuned, it was inferred via the
    $\temaopt$ scaling law (\cref{sec:lambda}).  Additional sweeps
    of $\eta$ were done at each $B$ at 610M-20TPP scale for the
    experiments in \cref{sec:tema_results}. Around 400 different LLMs
    were trained in total across all the
    experiments.\label{tab:train_steps}}
\begin{tabular}{@{}ccccc@{}}
\toprule
Model & TPP  & $D$ & Number of $B$ & Number of B with $\lambda$ tuned \\ \midrule
111M  & 20   & 2.19B & 8 & 8 \\
111M  & 80  & 8.76B & 8 & 8 \\
111M  & 200  & 21.9B & 7 & 7 \\
111M  & 320  & 35.0B & 8 & 8 \\
111M  & 1280  & 140.1B & 6 & 1 \\
266M  & 20   & 5.31B & 7 & 7 \\
266M  & 80   & 21.2B & 7 & 7 \\
266M  & 320   & 85.0B & 6 & 6 \\
266M  & 1280   & 339.8B & 1 & 0 \\
610M  & 20  & 12.1B & 8 & 8  \\
610M  & 80 & 48.5B & 7 & 7   \\
610M  & 200 & 121.3B & 6 & 6 \\
610M  & 320 & 194.1B & 2 & 1   \\
1.7B  & 20  & 34.3B & 7 & 7   \\
1.7B  & 80  & 137.2B & 7 & 1   \\
1.7B  & 320  & 548.6B & 1 & 0   \\
3.3B  & 20  & 66.5B & 1 & 0   \\
3.3B  & 23  & 76.5B & 1 & 0   \\
3.3B  & 30  & 99.8B & 2 & 1   \\ \bottomrule
\end{tabular}
\end{table}

\cref{tab:model_info} provides details on the model architecture and
hyperparameters for models used in the experiments.
\cref{tab:train_steps} provides, for each model scale and TPP, the
dataset sizes used in training, the number of batch sizes tested, and
the number of batch sizes for which $\lambda$ was tuned.
Around 400 models in total were trained for the main experiments.

All the models in our main experiments were trained on the SlimPajama
dataset~\citep{cerebras2023slimpajama}, a cleaned and deduplicated
version of the RedPajama dataset.  We use the
GPT-2~\citep{radford2019gpt2} vocabulary of size 50257, and a context
length of 2048 tokens.
Following standard practice, we do not apply weight decay or bias to
LayerNorm layers.
AdamW settings are $\beta_1 = 0.9$, $\beta_2 = 0.95$, and $\epsilon = 1$e$-8$.
Validation loss is always computed over a held-out 1.1B tokens,
regardless of training TPP\@.  We report cross-entropy loss.
By default we parameterize with $\mup$, with hyperparameters set via
proxy tuning, as described below.

For a given TPP, all models have the exact same warmup phase: a linear
warmup of the learning rate from 0 to the maximum value.
In all our runs, warmup was 10\% of the total steps.
Learning rate warmup is standard practice in LLM
pre-training~\citep{brown2020language,rae2022scaling,biderman2023pythia,dubey2024llama,kosson2024analyzing}.

All models in the main experiments were trained on a Cerebras CS-3
system.  610M-parameter 20TPP models take roughly 6 hours each to
train on a single CS-3.

For a given model configuration, we find results to be very stable
across random seeds.  To quantify the variance, we repeated
111M-parameter, 20~TPP training four additional times for six
different hyperparameter settings, resulting in 5 total validation
loss results for each of the six training runs.  Standard deviation of
the validation loss was below 0.003 in all cases.

\paragraph{Proxy model hyperparameter tuning}\label{sec:proxy_tuning}

\begin{table}
    \centering
    \caption{Tuned hyperparameters for $\mup$ proxy model\label{tab:mup_hps}}
    \begin{tabular}{cc}
         \toprule
         $\sigma_{W,\text{base}}$& $8.67$e-$02$ \\
         $\hateta$& $\maxlrdetail$\\
         $\alpha_{\text{input}}$& $9.17$\\
         $\alpha_{\text{output}}$& $1.095$\\
         \bottomrule
    \end{tabular}
\end{table}

To find the optimal $\mup$ hyperparameters (HPs), we trained a 39M
proxy model using a width $d_{\text{model}}$ of 256, with 24 layers
and head size of 64.  We trained this model on 800M tokens with a
batch size of 256 sequences and a context length 2048.  We randomly
sampled 350 configurations of base learning rates, base initialization
standard deviation, and embedding and output logits scaling factors,
and used the top-performing values as our tuned HPs
(\cref{tab:mup_hps}).

\section{Additional related work}\label{sec:related}
\subsection{Optimizers for large-batch training}

Prior work has explored optimizers designed specifically for
large-batch training, including LARS~\citep{you2017large} and
LAMB~\citep{you2019large}.  It is instructive to consider these prior
findings in light of the scaling laws from our paper.
In particular, both original BERT~\citep{devlin2019bert} and the LAMB
replication were trained on 85.2B tokens.  Applying our fitted
$\bcrit$ power law over $D$=85.2B, we obtain an estimated $\bcrit$ of
about 12M tokens.  Original BERT was trained for 90\% of steps with a
batch size of 65K tokens (512 sequences of length 128).  LAMB
increased the batch size to 4M tokens (32K sequences), justifying
their claim, ``BERT training can be reduced from 3 days to just 76
minutes'' \citep{you2019large}.  However, based on the predicted
$\bcrit$ of 12M, batch size 4M is still well within the expected range
of efficient batch sizes. Moreover, the LAMB paper later notes, ``we
did not observe any speedup by increasing the batch size from 65536 to
131072 [sequences, or 16.8M tokens].''  In other words, they reach the
point of diminishing return \emph{exactly where $B$ exceeds our
predicted $\bcrit$}.

It is likely that some optimization issues solved by LAMB (to enable
stable large-batch training) are solved other ways in modern LLM
training setups, via, e.g., gradient clipping, pre-LayerNorm
placement, better initialization and stability control through $\mup$,
etc.  Scaling $\lambda$ rather than $\eta$ with $B$, as we propose,
further supports stable, efficient training. However, gradient
redundancy imposes an inherent limit on useful batch sizes, ensuring
critical batch size remains relevant.

\subsection{Critical batch size}\label{sec:app_bcrit}

Observations of critical batch size have previously been related to
data complexity~\citep{golmant2018computational}, loss
curvature~\citep{ma2018power,zhang2019algorithmic}, and model
architecture~\citep{shallue2019measuring}.

\citet{merrill2025critical} define $\bcrit$ as the largest $B$ such
that loss does not degrade by more than a fixed fraction $\epsilon$
from the $\bopt$ setting.  They measure this $\bcrit$ instantaneously
throughout training, by repeatedly branching from a checkpoint with
different $B$ settings and assessing the impact on loss.

Recent work also defines $\cbs$ in terms of how $\eta$ scales with
$B$~\citep{filatov2024time,li2024surge}; unlike our work, these recent
studies use a constant learning rate schedule and no weight decay.

We follow \citet{mccandlish2018empirical}'s definition of $\bcrit$
(\cref{def:cbs}).
Given various theoretical assumptions, \citet{mccandlish2018empirical}
derived a direct equivalence between $\bcrit$ and what they call the
\emph{gradient noise scale} (GNS): the variation of the gradients
between different training examples.
However, they noted that the GNS ``accurately predicts the largest
usable batch size (at the order of magnitude level),'' which is below
the level of precision needed for large-scale training.
\citet{merrill2025critical} recently found ``the gradient noise scale
underestimates the CBS [i.e., $\bcrit$].''
This lack of precision may be why, in Kaplan et al's original scaling
laws paper~\citep{kaplan2020scaling}, they note that, ``although the
critical batch size roughly matches the gradient noise scale, we are
using a direct [empirical] measurement of $\bcrit$.'' Our approach to
measuring $\bcrit$ (\cref{sec:estimating_bcrit}) similarly provides a
direct empirical measurement, but one that can be efficiently computed
with any learning rate schedule or optimizer.

\subsection{Detailed comparison with \citet{zhang2024how}}\label{sec:appendix_zhang}

Here we provide further comparison with the concurrent work
by \citet{zhang2024how}.
The primary point of distinction of our paper is that we conducted a
large-scale empirical study into the scaling of AdamW's weight decay
hyperparameter (including its scaling with $B$), ultimately deriving a
precise power law for the optimal AdamW timescale in
tokens-per-parameter.  \citet{zhang2024how} did not use weight decay.
Further, we also explored scaling of $\bopt$ in addition to $\bcrit$.
Beyond use of weight decay, further methodological differences in our
main experiments include that we used a longer context length (2048
vs.\ 512), a cleaner dataset (SlimPajama vs. C4), the $\mup$
parameterization, a decaying LR schedule (more on this below), and
that we tuned HPs at most $N$, $D$, $B$ (\cref{tab:train_steps}),
while \citet{zhang2024how} performed a HP sweep for a 151M model, and
re-used optimal values at other scales.
We now focus on differences in estimating and measuring the scaling of
$\bcrit$.

\paragraph{Estimating $\bcrit$ for a specific target loss}

Both our work and \citet{zhang2024how} require measuring, for
different batch sizes, how many training steps it takes to reach a
particular target loss.
Since the number of steps to reach that loss is not known \emph{a
priori}, it is inherently difficult to study $\bcrit$ when using a LR
decay schedule, where you must specify the number of steps in advance.
Using a constant LR (as was done in early work on
$\bcrit$~\citep{mccandlish2018empirical}) simply does not result in
competitive models~\citep{bergsma2025straight}.
Unfortunately, it is not feasible to \emph{search} for the precise
step count needed, i.e., by conducting full training runs with
different schedules/step budgets.

\citet{zhang2024how} creatively solve this issue by conducting a
single training run at a constant LR, while using weight averaging to
generate higher-quality checkpoints for evaluation.  With this
approach, they still ``need to frequently evaluate the model on a
holdout evaluation set'' \citep{zhang2024how}.

Given LR decay, as opposed to weight averaging, remains the standard
practice for current state-of-the-art LLMs, we independently developed
a different approach.  This led to the novel method described in our
paper.  In contrast with~\citet{zhang2024how}, we do not need to
frequently evaluate the model, as we instead fit a $B$-specific loss
power law through a few validation loss values
(\cref{sec:estimating_bcrit}).  Indeed, our approach may improve the
efficiency of \citet{zhang2024how}'s method, as it would obviate the
cost of continuous validation, which concerned them (see their section
``Evaluation data size and frequency'').

\paragraph{Estimating the $\bcrit$ power law}

Collecting $\bcrit$ data across multiple model scales and loss targets
is expensive.  \citet{zhang2024how} establish $\bcrit$ scaling in $D$
through three targeted experiments:
\begin{itemize}
\item measuring $\bcrit$ while scaling $N$ but leaving $D$ fixed to 3.07B
\item measuring $\bcrit$ while scaling $D$ but leaving $N$ fixed to 302M
\item measuring $\bcrit$ while scaling both $N$ and $D$ proportionally (at 20 TPP)
\end{itemize}
Interestingly, $\bcrit$ was found to only scale weakly in $N$, but
scale similarly whenever $D$ is scaled.  They then fit a power law to
their data points for the 302M-parameter model, obtaining the fit
\mbox{$\bcrit = 22.91 D^{0.47}$} (in tokens).

In comparison, we took a more brute-force approach, computing $\bcrit$
across many different $N$ and $D$ values, and ultimately fitting our
$\bcrit$ power law across multiple different model sizes and TPP
settings (\cref{fig:hook}, $\rightfig$).  Also,
unlike~\citet{zhang2024how}, we assessed the quality of fit via
computation of both $\Rtwo$ values and parameter quantiles via
bootstrapping (\cref{sec:results_cbs}).

Recall also that \citet{zhang2024how} used a different definition of
critical batch size.  Let us denote their quantity $\cbszhang$.  They
set $\cbszhang$ to be the $B$ such that the data required to reach a
loss target is $1.2 \times \dmin$ (i.e., $1.2 \times$ the data
required with $\bopt$).
%Their fitted power law is
%\mbox{$\cbszhang = 22.91 D^{0.47}$}, where $\cbszhang$ is in tokens.

We can use \cref{eq:s_extra} to align their fitted law with our own.
By this equation, we have:
\begin{align*}
  D &= \dmin(1 + \frac{\cbszhang}{\cbs}) \\
    &:= \dmin(1.2) \\
  \Rightarrow \frac{\cbszhang}{\cbs} &= 0.2 \\
  \Rightarrow \cbs &= 5 \cbszhang
\end{align*}
Thus, to convert their coefficient to our scale, we multiply it by 5,
and, dividing by the number of tokens in our sequences, obtain
$\cbszhang = 0.0559 D^{0.47}$.
The $\cbszhang$ coefficient (0.0559) is 19\% larger than our own
(0.0471), perhaps reflecting differences in training setup or data
quality (and worth investigating further in future work).
However, the exponents are quite similar (0.47 vs. 0.462), suggesting
that both works are independently measuring the same fundamental
scaling behavior.

We emphasize that $\bcrit$ directly reflects the fundamental limit to
data parallelism in training neural networks.  Given the significant
implications of $\bcrit$ scaling in $D$ rather than $C$ or $L$
(including those discussed in \cref{sec:balancing}), we note the
scientific and practical value in having different approaches
independently observe this same phenomenon.

\subsection{Hyperparameter scaling with $B$}

It has long been recognized that the optimal learning rate, $\etaopt$, scales
with $B$, with reports of both
linear~\citep{krizhevsky2014one,chen2016revisiting,smith2017bayesian,smith2018dont}
and square-root
scaling~\citep{hoffer2017train,you2019large,malladi2022sdes}.
Recent work has found $\etaopt$ to \emph{decrease} when $B
> \cbs$~\citep{li2024surge,filatov2024time}, which resonates with our
own findings (\cref{fig:tepochs}, $\rightfig$).
%
%\citet{li2024surge} look at learning rate scaling with Adam, they see an
%increase, then a decrease, then an increase again, but at many
%multiples of $\cbs$...
%- That Time Transfer paper also sees this with MUP
%
Since it is difficult to predict exactly how $\eta$ will scale with
$B$, studies of $\cbs$ have often done full HP sweeps at each
$B$~\citep{mccandlish2018empirical,shallue2019measuring}.
% \citet{zhang2024how} perform a HP sweep for a 151M model, and then
%re-use optimal values at other scales.

The only work we are aware of that specifically recommends scaling
weight decay with $B$ is \citet{loshchilov2017decoupled}, who suggest
$\lambda \propto \sqrt{B}$, though this rule is not evaluated
systematically. It is also important to note that
\citet{loshchilov2017decoupled} use the \emph{independent} form of
weight decay, where decay is applied independently of the learning
rate $\eta$, unlike common implementations such as AdamW in
PyTorch~\citep{wortsman2023small}. In these more typical
\emph{dependent} implementations, weight decay is scaled by $\eta$, so
any increase in $\eta$ with $B$ (e.g., $\eta \propto B$ or $\sqrt{B}$)
already increases the effective weight decay strength accordingly.

%399R: Note in Loshchilov & Hutter (2017) they use fully decoupled weight decay. Scaling it with the sqrt(B) may therefore simply be equivalent to scaling the learning rate by sqrt(B) in the more common not-fully-decoupled AdamW implementation (default in PyTorch).

\subsection{$\tepoch$ and effective learning rates}

The concept of effective learning rates, influenced by weight decay,
has been widely
discussed~\citep{van2017l2,hoffer2018norm,zhang2018three,chiley2019online,li2019exponential,wan2020spherical,kosson2023rotational,dangelo2024why}.
In its simplest form, the effective or \emph{intrinsic} LR is simply
$\eta\lambda$, but in these prior works, effective LRs typically
measure functional updates relative to weight magnitude, which is
particularly relevant for normalization-based networks.
Comparison of the effects of $\lambda$ vs.\ $\eta$ adjustments in the
context of LR decay schedules was explored
in~\citet{bergsma2025straight}.

The behavior of effective LRs (relative update sizes) over the course
of training has been studied comprehensively
by \citet{kosson2023rotational}, including comparing the effects of
increasing $\eta$ vs.\ increasing $\lambda$.
This work shows that higher $\eta$ values can cause large relative
updates early in training, which can destabilize training or require
longer warmups~\citep{kosson2024analyzing}.
High $\eta$ and low $\lambda$ can also lead to larger weight
norms~\citep{kosson2023rotational,dangelo2024why}, which also has a
destabilizing effect, particularly on low-precision training.
These effects may explain why we were able to achieve higher effective
LRs $\eta\lambda$ by tuning $\lambda$ rather than tuning $\eta$ with
$B$ (\cref{fig:tepochs}, $\middlefig$).

For a given dataset size $D$, the $\tepoch$ and the batch-normalized
effective LR $\frac{\eta\lambda}{B}$ are equivalent, and thus
effective LRs and the AdamW timescale can be viewed as different
perspectives on the AdamW optimization process.

\section{Scaling of $\tepoch$ and $\lambda$: additional details and results}

%\input{figures/fig_more_tepoch_scaling.tex}

%In TODO, we plot the optimal $\tepoch$ as a function of batch size for
%all of the model scales and TPP levels where we did hyperparameter
%sweeps.  The optimal $\tepoch$ is taken from a coarse grid (as
%$\lambda$ is swept by powers of two), so when the true $\tepochopt$ is
%between two grid points, the sampled optimum may oscillate, as we see
%at smaller model scales.  The optimal $\tepoch$ stays around a common
%value increasing when $B > \cbs$.  Note that the point of increase
%tends to scale both with TPP and model scales, as expected if $\cbs$
%is a power law in $D$ (as, at a fixed TPP, an increase in model scale
%also requires an increase in $D$).

\subsection{$\lambda$ scaling with $B$}

\begin{figure*}[t]
  \centering
  \begin{subfigure}[b]{\appendixwidth\textwidth}
    \centering
    \includegraphics[trim={0.3cm 0.4cm 0.3cm 0.3cm}, clip, width=\linewidth]{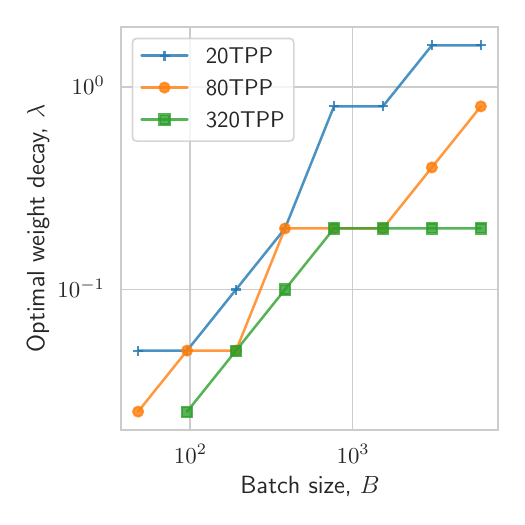}
    \caption{111M scale}
    \label{fig:more_wd_scaling_111M}
  \end{subfigure}
  \hfill %
  \begin{subfigure}[b]{\appendixwidth\textwidth}
    \centering
    \includegraphics[trim={0.3cm 0.4cm 0.3cm 0.3cm}, clip, width=\linewidth]{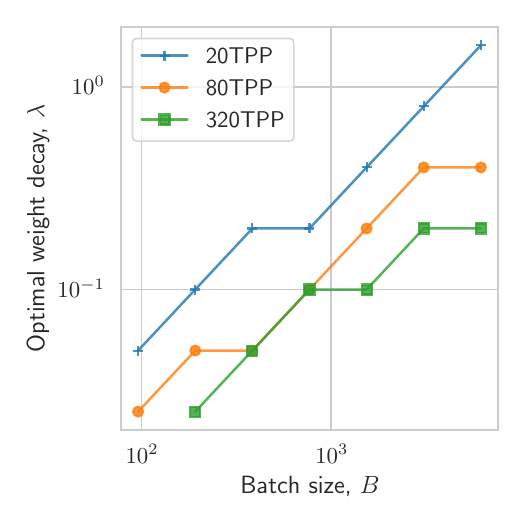}
    \caption{266M scale}
    \label{fig:more_wd_scaling_266M}
  \end{subfigure}
  \begin{subfigure}[b]{\appendixwidth\textwidth}
    \centering
    \includegraphics[trim={0.3cm 0.4cm 0.3cm 0.3cm}, clip, width=\linewidth]{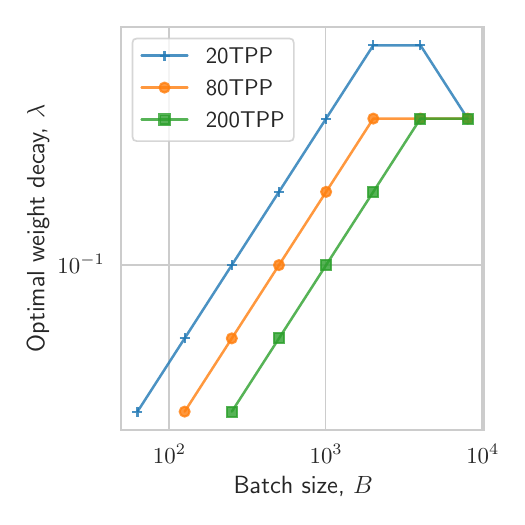}
    \caption{610M scale}
    \label{fig:more_wd_scaling_617M}
  \end{subfigure}
  \hfill %
  \begin{subfigure}[b]{\appendixwidth\textwidth}
    \centering
    \includegraphics[trim={0.3cm 0.4cm 0.3cm 0.3cm}, clip, width=\linewidth]{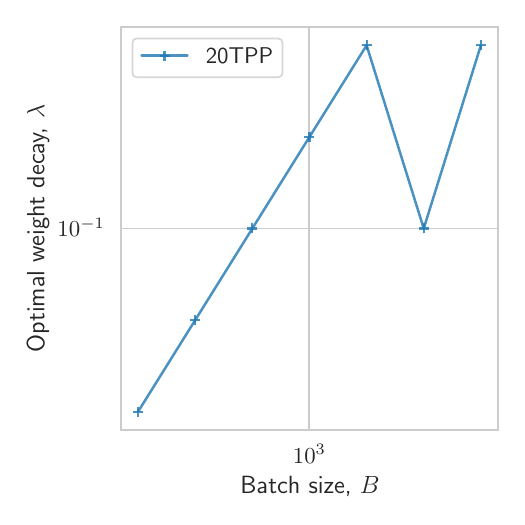}
    \caption{1.7B scale}
    \label{fig:more_wd_scaling_1_7B}
  \end{subfigure}
  \caption{\textbf{Optimal weight decay scaling with $B$}: The optimal
    weight decay increases linearly over small batch sizes---until
    $B > \cbs$.
    \label{fig:more_wd_scaling}}
\end{figure*}

\cref{fig:more_wd_scaling} shows how optimal $\lambda$ changes across
$B$, for all of the model scales and TPP levels where we did
hyperparameter sweeps.
There is a strong linear relationship between $\lambda$ and $B$ over
the smaller batch sizes $B < \bcrit$, with optimal $\lambda$
eventually plateauing (or decreasing).
Note the standard use of
$\lambda$=0.1~\citep{hoffmann2022empirical,brown2020language,almazrouei2023falcon,alephalpha2024introducing}
is only optimal at specific $B$, and this $B$ changes with TPP\@.

% This agrees with recent work showing that $\eta$ decreases when $B >
%\cbs$, when using the Adam
%optimizer~\citep{li2024surge,filatov2024time}.  Could the transition
%between linear scaling and no-further-scaling really be what's
%driving the square-root scaling rules discussed in related work?

\subsection{Additional details on $\tepoch$ fitting}\label{sec:appendix_tepoch_fitting}

\renewcommand{\algorithmiccomment}[1]{$\triangleright$ #1}
\begin{algorithm}[t]
  \caption{Generating the optimal $\tepoch$ power law}
  \label{alg:tepoch_fitting}
  \begin{algorithmic}
    \small
    \STATE {\bfseries Input:} small batch size $B$, optimal per-$N$ learning rates $\eta$ 
    \STATE Initialize $\tepochPoints = [$ $]$
    \FOR{$N$ {\bfseries in} $\modelScales$}  % N
      \FOR{$D$ {\bfseries in} $10N, 20N, 80N, 320N, ...$} % D
        \STATE Reset $\lossPoints = [$ $]$
        \FOR{$\lambda$ {\bfseries in} $\lambdaRange$} % lambda
          \STATE Train $\LLM(N,D,B,\lambda,\eta)$, get validation loss $L'$
          \STATE $\tepoch = \nicefrac{B}{\eta \lambda D}$
          \STATE $\lossPoints[\tepoch] = L'$
        \ENDFOR % L
        \STATE $\tepochopt = \argmin_{\tepoch}(\lossPoints)$
        \STATE $\tepochPoints.add( \langle \tpp$=$\nicefrac{D}{N}, \tepochopt \rangle )$
      \ENDFOR % D
    \ENDFOR % N
    \STATE Fit $\ctepochalg$, $\mtepochalg$ for $\tepochopt = {\ctepochalg} \tpp^{\mtepochalg}$ on $\tepochPoints$
  \end{algorithmic}
\end{algorithm}

We now describe how we obtained the optimal $\tema$ values at specific
model scales and TPP ratios.
Rather than taking the empirical minimum loss, we fit a parabola to
the $\langle \mathit{L}, \tepoch \rangle$ points in log space and took
the analytic minimum of the parabola.  If we have multiple loss values
at the same $\tepoch$ (e.g., our data for a single scale and TPP
comprises multiple different batch sizes), we only kept the lowest
loss points at each $\tepoch$ prior to parabola fitting.
We used validation loss on the held-out validation set.
For our $\tepoch$ calculations, we input $B$ in units of \emph{tokens}
(in contrast to the \emph{output} of our reported $\bopt$ and $\bcrit$
scaling laws, which we report in units of sequences, of 2048 tokens).
\cref{alg:tepoch_fitting} sketches the full procedure for generating
the $\tepoch$ power law (\cref{eq:tepoch_scaling}).
%
% Implied by taking the best ones?
%Note we are collecting $\tepoch$ here for runs where $B < \cbs$.

\subsection{Relationship to prior power laws}

\subsubsection{Relationship to $\etaopt$ scaling laws in dataset size, $D$}\label{sec:eta_scaling}

Both \citet{shen2024power} and \citet{bjorck2024scaling} propose
scaling laws for the optimal learning rate, $\etaopt$, as a power law
in the amount of data, $D$:
\begin{equation*}
  \etaopt = B \cdot \cetad \cdot D^{\metad}
\end{equation*}
We now discuss how this power law also follows from the power law of
$\tepochopt$ in TPP\@.  By \cref{eq:tepoch_scaling}, we have:
\begin{align*}
  \tepochopt(\tpp) &= {\ctepoch} \cdot \tpp^{\mtepoch} \\
  &= {\ctepoch} \cdot \left( \frac{D}{N} \right)^{\mtepoch}
\end{align*}
Substituting in the definition of $\tepoch$ (\cref{eq:tepoch}), and
assuming $\lambda$, $B$, and $N$ are
fixed,\footnote{\citet{bjorck2024scaling}\ use a fixed
$\lambda$=$0.1$, a fixed batch size of 0.5M tokens (for most of their
experiments), and fit scaling laws separately at different model sizes
(except in their Section~4).  \citet{shen2024power}\ use $\mup$ to
adjust the LR for different model scales, so the derivation applies at
any particular $N$; they do not report which optimizer is used nor any
of its settings.} this implies $\etaopt$ will scale as:
\begin{align}\label{eq:etaopt_equivalence}
  \frac{B}{\etaopt\lambda D} &= {\ctepoch} \cdot \frac{D^{\mtepoch}}{N^{\mtepoch}} \notag \\
  \Rightarrow \etaopt &= B \left( \frac{N^{\mtepoch}}{\lambda \cdot \ctepoch} \right) D^{-(\mtepoch + 1)} \notag \\
  &= B \cdot \cetad \cdot D^{\metad}
\end{align}
\begin{equation*}
  \text{where} \quad 
  \cetad = \frac{N^{\mtepoch}}{\lambda \cdot \ctepoch}
  \quad \text{and} \quad
  \metad = -(\mtepoch + 1)
\end{equation*}
\cref{eq:etaopt_equivalence} is exactly the form of the power law used
in \citet{shen2024power}, and explains the results across batch sizes
seen in \citet{bjorck2024scaling}, as discussed in
\cref{sec:tema_results}.

\paragraph{Comparison to fit in Power Scheduler~\citep{shen2024power}}\label{sec:appendix_power_scheduler}

Given $\cetad = \frac{N^{\mtepoch}}{\lambda \cdot \ctepoch}$ and $\metad =
-(\mtepoch + 1)$, we can use these formulas to compare our fit
coefficients to those in \citet{shen2024power}.

In \citet{shen2024power}, they find $\metad = -0.51$.
In our case, $\mtepoch = -0.520$, and therefore $\metad = -0.48$, which
is quite similar.

Comparing our $\cetad$ to their $\cetad$ ($= 4.6$) is a bit less
straightforward.  First of all, \citet{shen2024power} inputs $B$
in sequences (of length 4096).  We thus convert to the scale of our
coefficient by dividing by their sequence length, obtaining $\cetad =
0.0011$.
Secondly, the power law of \citet{shen2024power}\ is actually for
the \emph{base} $\mup$ learning rate $\hateta$, while our derivation
above assumes the adjusted (final) learning rate $\eta$
(\cref{sec:mup}).

Let us first compare $\cetad$ coefficients at the proxy-model scale,
i.e., where $\eta = \hateta$.
If we were to use a 28M-parameter proxy model, and a default
$\lambda$=0.1 (and using our fit values of $\ctepoch=1.084$ and
$\mtepoch=-0.527$), then, by $\cetad = \frac{N^{\mtepoch}}{\lambda
  \ctepoch}$, our $\cetad$ would also equal $0.0011$.

Now we consider how our coefficient varies when $N$ scales.  To
convert our $\eta$ scaling law into one for the base $\hateta$, we can
instead use $\cetad = \frac{N^{\mtepoch}}{\lambda \cdot \rho \cdot \ctepoch}$, where $\rho = P/W$, $P$ is the width of the proxy model,
and $W$ is the width of the target model.
The width also affects the number of parameters, $N$, and hence the
term $N^{\mtepoch}$.  In Transformers, $N$ scales roughly as $N
\propto L W^2$, where $L$ is the model depth and $W$ is the model
width.  If we round the fitted exponent to $\mtepoch \approx -0.5$,
and substitute the value $\rho \propto 1/W$ into the denominator, we
therefore have:
\begin{align*}
  \cetad &= \frac{N^{\mtepoch}}{\lambda \rho \ctepoch} \\
  &\propto \frac{N^{-0.5}}{\frac{1}{W}} \\
  &\propto {(L W^2)}^{-0.5}W \\
  &\propto {L}^{-0.5}{(W^2)}^{-0.5}W \\
  &\propto {L}^{-0.5}
\end{align*}
which is invariant to changes in $W$---i.e., the $\mup$ adjustment
cancels out the model scaling in width.
So, if we only scale $W$, $\tema$ scaling would stay in agreement with
the Power Scheduler recipe, but if we increase depth, $\tema$ scaling
would decrease $\eta$ proportional to $\nicefrac{1}{\sqrt{L}}$ in a
manner that is not accounted for in \citet{shen2024power}.

The key point is that the scaling law used by
\citet{shen2024power}\ is valid, indeed, has similar fitted exponents,
to what would be predicted by the $\tepochopt$ scaling law---but in a
specific context only (small models, or models only scaling in width).
Moreover, we have shown it may be less effective to adjust $\eta$ in
order to optimize $\tepoch$ (as these approaches implicitly do); we
obtained superior results by instead adjusting $\lambda$.  By
considering $\eta$, $\lambda$, and $B$ holistically, our scaling laws
are a superset of these laws for $\etaopt$, as well as other laws that
we discuss further presently.

\subsubsection{Relationship to $\etaopt$ scaling laws in model size, $N$}\label{sec:appendix_etaopt}

\cref{sec:tema_methods} gave our recipe for tuning hyperparameters,
for an arbitrary $N$, $D$, and $B$ setting.  Here we advocated setting
peak $\eta$ to the $\mup$-adjusted learning rate (where the base
learning rate comes from proxy-tuning).
Rather than further adjusting this LR based on the dataset size or
batch size, we argued for instead adjusting $\lambda$ so that
$\tepoch$ is tuned to its optimal value.  Based on both theory, and
our empirical findings comparing tuning $\eta$ to tuning $\lambda$, we
believe that using $\mup$ to scale $\etaopt$ with model width is
sufficient for well-tuned models.
That is, the \emph{theoretical} scaling law for $\etaopt$ (in model
width), given by $\mup$, is sufficient for good performance.
We discuss this perspective further in this section, specifically how
the $\mup$ scaling law can explain recent work in \emph{empirical}
$\etaopt$ power laws.

As noted in \cref{sec:mup}, when using $\mup$, a base $\eta$ is tuned
on a small proxy model, and then scaled depending on the width of the
target model.  Let $W$ be the width of the target model, and let $P$
be the width of the proxy model.  $\mup$ prescribes scaling the
optimal base learning rate, $\hatetaopt$, down to $\etaopt =
\rho\hatetaopt$, where $\rho=P/W$.  That is, $\etaopt =
P\hatetaopt/W$.
As models grow in size, $P$ and $\hatetaopt$ do not change, so
$\etaopt$ will scale $\propto 1/W$.
\citet[Figure~2]{dey2024practitioner} show that, indeed, a range of
LLMs from the GPT, Llama, and DeepMind series are roughly following a
scaling law where their chosen learning rate, $\eta$ is following
$\eta \propto {1}/{W}$.
In other words, if one were to build a scaling law for $\etaopt$ based
on published LLM settings, it would roughly obey the $\mup$
theoretical scaling law.

Furthermore, we can develop a scaling law for $\etaopt$ in model size,
$N$, using $\mup$, and show that it matches a recent empirical scaling
law by \citet{porian2024resolving}.  The number of model parameters in
any Transformer-based LLM scales roughly in the depth, $L$, and width,
$W$, as $N \propto L W^2$.  If we assume that we maintain a fixed
width-to-depth ratio, i.e., $R = W/L$, or $L = W/R$, then we have $N
\propto W^3$, or $W \propto N^{1/3}$.
Now, since $\mup$ prescribe scaling $\etaopt \propto W^{-1}$, then for
a fixed aspect ratio, $\etaopt \propto N^{-1/3}$.

Taking a very different approach, \citet{porian2024resolving}
developed an empirical scaling law for $\etaopt$ as a function of the
number of model parameters.  At each model scale, they trained with a
variety of batch sizes and learning rates, and found the optimal
settings of these hyperparameters.  All models were trained to 20
TPP\@.  They then fit a power law through the optimal LR settings, and
found that $\etaopt \propto N^{-1/3}$, exactly as would be expected if
one simply followed $\mup$.

As it provides a principled approach to scaling hyperparameters,
$\mup$ can adapt to scaling when aspect ratio is not fixed.  We
therefore advocate using $\mup$ to set $\etaopt$, rather than fitting
special $\etaopt$ power laws.
However, with regards to our overall approach, it does not actually
matter whether one uses the $\mup$ theoretical scaling law \emph{or}
an empirical one.  The key point is that these laws can be used to set
$\eta$ at a particular model scale, while the $\tepoch$ law should
further be used to set $\lambda$ depending on the $B$ or $D$ values.

%Both \citet{porian2024resolving} and \citet{bi2024deepseek} utilize
%joint scaling laws for $\eta$ and $B$.  The benefit of this approach
%is that $\eta$ can be tuned for early-stage training dynamics
%(maximizing movement from initial conditions, e.g., via $\mup$, the
%\emph{maximal} update parameterization), while $B$ simultaneously
%optimizes $\tepoch$.  The drawback, as mentioned in
%Section~\ref{sec:related} when discussing power laws for $\bopt$, is
%that if we rely on tuning $B$ to optimize $\tepoch$, we do not have
%flexibility to pick a batch size that optimizes the tradeoff between
%parallelism and compute efficiency.  We discuss batch size laws
%further in the following section.

\subsection{The EMA perspective and learning rate schedules}\label{sec:ema_lr}

To understand how the EMA view applies with a dynamic LR schedule, we
follow the discussion of \citet{bergsma2025straight}, who extended the
formulation in \citet{wang2024how}.
\citet{bergsma2025straight} consider EMAs with time-varying smoothing,
$\alpha_t \in [0, 1]$.  Letting $\alpha_1=1$ (i.e., $y_1=x_1$), they
express $y_t$ in terms of all inputs $x_t$:
\begin{eqnarray}\label{eqn:extended_ema}
  y_1 &=& \alpha_1 x_1  \notag, \\
  y_2 &=& (1 - \alpha_2) \alpha_1 x_1 + \alpha_2 x_2, \cdots  \notag \\
%  y_3 &=& (1 - \alpha_3) (1 - \alpha_2) \alpha_1 x_1 + (1 - \alpha_3) \alpha_2 x_2 + \alpha_3 x_3, \cdots \notag \\
%  &\cdots&  \notag \\
  y_t &=& \sum_{i=1}^t \left( \prod_{j=i+1}^{t} (1 - \alpha_j) \right) \alpha_i x_i
\end{eqnarray}

The EMA coefficient on each input is denoted $c_{t,i}$, where
\mbox{$c_{t,i} = \left( \prod_{j=i+1}^{t} (1 - \alpha_j) \right)
  \alpha_i$}.
In other words, $c_{t,i}$ reflects the contribution of input $x_i$ to
output $y_t$ at time $t$, such that
$y_t = \sum_{i=1}^t c_{t,i} x_i$.
Unlike a standard EMA with a fixed smoothing parameter, in this
extended EMA the coefficients need not decrease exponentially as $i$
decreases.  Indeed, any set of coefficients can be generated by some
particular smoothing schedule.

\begin{figure*}[t]
  \centering
  \begin{subfigure}[b]{0.32\textwidth}
    \centering
    \includegraphics[trim={0.3cm 0.4cm 0.3cm 0.3cm}, clip, width=\linewidth]{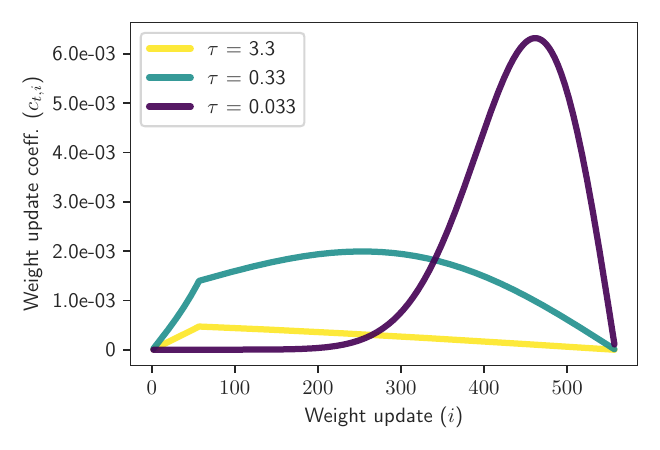}
    \label{fig:emas_tepoch_2TPP}
    \caption{557 steps}
  \end{subfigure}
  \hfill
  \begin{subfigure}[b]{0.32\textwidth}
    \centering
    \includegraphics[trim={0.3cm 0.4cm 0.3cm 0.3cm}, clip, width=\linewidth]{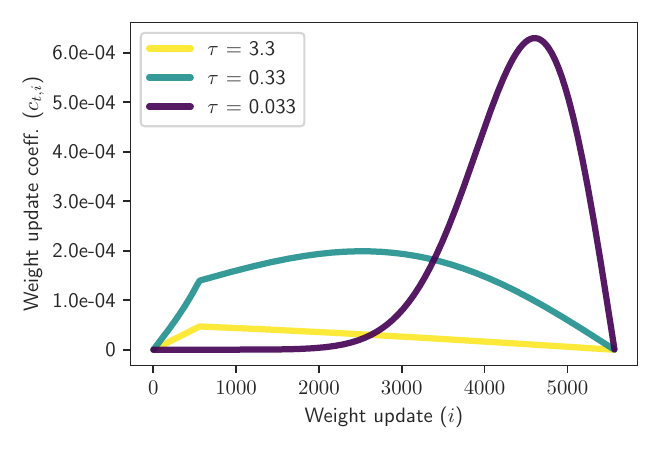}
    \label{fig:emas_tepoch_20TPP}
    \caption{5568 steps}
  \end{subfigure}
  \hfill 
  \begin{subfigure}[b]{0.32\textwidth}
    \centering
    \includegraphics[trim={0.3cm 0.4cm 0.3cm 0.3cm}, clip, width=\linewidth]{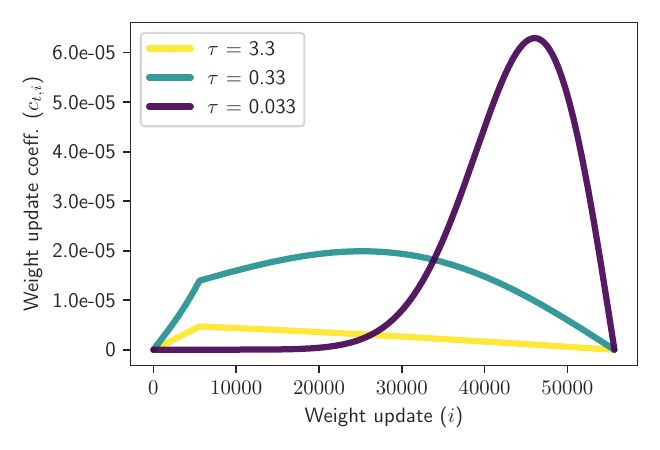}
    \label{fig:emas_tepoch_200TPP}
    \caption{55680 steps}
  \end{subfigure}
  \caption{\textbf{$\tepoch$ is invariant to steps, with a learning
      rate decay schedule} (111M scale, proxy-tuned peak $\eta$ with
    linear decay-to-zero): Here we adjust weight decay, $\lambda$, in
    order to maintain $\tepoch$ at a constant value, decreasing
    $\lambda$ proportional to the increase in $S$.  Regardless of the
    total number of optimization steps, we see that the same $\tepoch$
    corresponds to the same \emph{shape} of the distribution of weight
    update coefficients (i.e., the same shape over the \emph{data},
    regardless of how the data is discretized in training).  For batch
    sizing, this means that if we use a constant $D$ but increase $B$
    by $K$$\times$ (decreasing $S$ by $K$$\times$), we will
    incorporate information across the {data} similarly---provided
    we use the same $\tepoch$.\label{fig:emas_tepoch}}
\end{figure*}

In terms of learning rate schedules for AdamW training, $\alpha_t =
\eta_t\lambda$ becomes the smoothing parameter at step $t$
(cf. \cref{sec:ema}).  The EMA itself, $y_t$, is the model parameters.
The EMA is over weight updates: a large coefficient $c_{t,i}$ means
the $i$th weight update contributes a lot to the EMA at step $t$.  The
EMA coefficients thus provide a more granular view of the contribution
timescale than $\tema$ alone.

We now study the question, how do the EMA coefficients change as the
step count changes?
We generated the $c_{t,i}$ coefficients for the linear decay-to-zero
LR schedule, and plot these coefficients at the final step (i.e.,
showing the contribution of weight updates to the final parameters).
We use the $\mup$-tuned and adjusted peak $\eta$, for 111M models.
The learning rate increases linearly to the peak for the first 10\% of
steps, then decreases from the peak to 0 for the remainder of steps.
We simulated three cases: where we take 557 steps, where we take 5568
steps, and where we take 55680 steps (5568 steps would be 20 TPP for a
111M model using $B$=192).
From the perspective of batch sizing, these different steps could be
achieved by decreasing the batch size twice by $10\times$.

We adjusted $\lambda$ for each step count in order to obtain the same
three specific $\tepoch$ values.  In \cref{fig:emas_tepoch}, we see
that the same $\tepoch$ implies the same \emph{shape} of coefficients
across the steps, and hence the same contribution over (normalized)
time.  That is, weight updates from the same portion of the training
data contribute equally to the final model parameters.\footnote{Note
we do not plot the initial coefficients $c_{t,0}$ here, but they are
equal across the scales for a given $\tepoch$, unlike the non-initial
coefficients, which reduce by $10\times$ as the number of steps
increases by $10\times$.  So a constant $\tepoch$ means both the same
contribution across the data, \emph{and} the same \emph{bias}
(dependence on initial weights).}

The key takeaway is that since $\tepoch$ is independent of the number
of steps, it theoretically provides a $B$-independent measure of the
AdamW timescale over weight updates, regardless of learning rate
schedule.
However, this equivalence for different $B$ breaks down when $B >
\cbs$ and weight updates themselves no longer contain linearly
$B\times$ the information of a single sample.

%\subsection{And EMA view of how $\tepoch$ changes with TPP}\label{sec:ema_tpp}
%
%TODO: draw the x-axis in ``proportion of steps'' and plot our fit
%values for TPP=5, TPP=10, TPP=20, TPP=80, TPP=320, TPP=1280.

\section{Scaling of $\bopt$ and $\bcrit$: additional details and results}

\subsection{Derivation of ``extra data'' \cref{eq:s_extra}}\label{sec:cbs_derivations}

% TOO COMPLICATED!
%It is worth noting that in \citet{mccandlish2018empirical}, the
%tradeoff equation (our \cref{eq:tradeoff}) is derived by averaging
%their Equation 2.7 over multiple optimization steps.  Their Equation
%2.7 gives the reduction in loss given a finite batch size of $B$ as a
%function of the maximum possible reduction in loss (with an infinite
%$B$) as 

%Something about how \cref{eq:tradeoff} itself arises from the
%per-step loss equation in~\citet{mccandlish2018empirical} (their
%equation 2.7).

\cref{eq:tradeoff} can be written as:
\begin{align*}
  \frac{D - \dmin}{\dmin} &= \frac{\smin}{S - \smin} \\
  \Rightarrow (D - \dmin)(S - \smin) &= \dmin \smin \\
  \Rightarrow D S - D \smin - S \dmin &= 0
\end{align*}
Given $B = {D}/{S}$, we can substitute in
$S={D}/{B}$ to get an equation in a single variable, from
which we can solve for $D$.
\begin{align*}
  \frac{D^2}{B} - D \smin - \frac{D \dmin}{B} &= 0 \\
  \Rightarrow D^2 - D B \smin - D \dmin &= 0 \\
  \Rightarrow D (D - B \smin - \dmin) &= 0 \\
  \Rightarrow D &= \dmin + B \smin
\end{align*}
Given $\cbs = {\dmin}/{\smin}$, we can substitute $\smin = {\dmin}/{\cbs}$ and obtain:
\begin{align*}
  \Rightarrow D &= \dmin + B \frac{\dmin}{\cbs} \\
  \Rightarrow D &= \dmin \left( 1 + \frac{B}{\cbs} \right)
\end{align*}

\subsection{Estimating $\cbs$}\label{sec:appendix_estimating_cbs}

We first provide some learnings from developing the $\cbs$ estimation
procedure.

First, regarding the functional form $L_{B}(D) = \irr + \dconst
D^{-\beta}$, we found that including the irreducible loss term $\irr$
was important for obtaining good fits.  $\irr$ conceptually represents
the Bayes risk \emph{plus} the minimum loss obtainable for a model of
size $N$ (i.e., the first two terms of \cref{eq:chinchilla}).
Second, only \emph{interpolated} points were reliable; we only compute
$\cbs$ for loss values where all points are between, or very near to,
curve fitting points.
Third, each power law should have at least $3$ points for fitting, in
order to capture the concavity of the scaling in $D$.
Finally, we sample our $B$ values logarithmically and, as in
\citet{mccandlish2018empirical}, perform our fits to
\cref{eq:tradeoff} in log space.

\begin{figure*}[t]
  \centering
  \begin{subfigure}[b]{\appendixwidth\textwidth}
    \centering
    \includegraphics[trim={0.3cm 0.4cm 0.3cm 0.3cm}, clip, width=\linewidth]{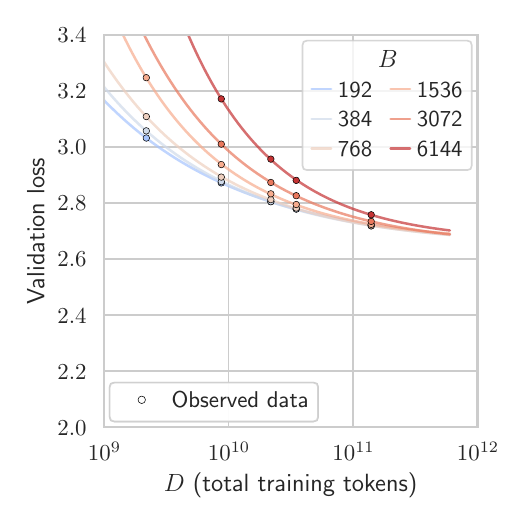}
    \caption{111M scale}
    \label{fig:quad_cbs_111M}
  \end{subfigure}
  \hfill
  \begin{subfigure}[b]{\appendixwidth\textwidth}
    \centering
    \includegraphics[trim={0.3cm 0.4cm 0.3cm 0.3cm}, clip, width=\linewidth]{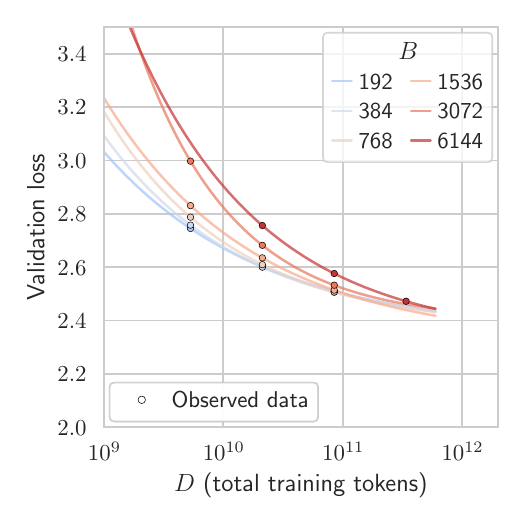}
    \caption{266M scale}
    \label{fig:quad_cbs_266M}
  \end{subfigure}
  % Next row
  \begin{subfigure}[b]{\appendixwidth\textwidth}
    \centering
    \includegraphics[trim={0.3cm 0.4cm 0.3cm 0.3cm}, clip, width=\linewidth]{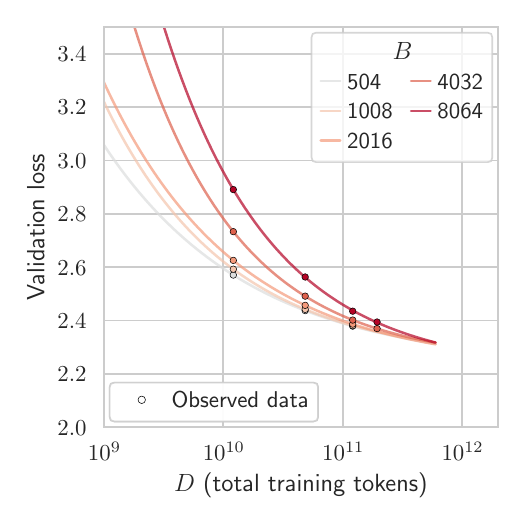}
    \caption{610M scale}
    \label{fig:quad_cbs_617M}
  \end{subfigure}
  \hfill
  \begin{subfigure}[b]{\appendixwidth\textwidth}
    \centering
    \includegraphics[trim={0.3cm 0.4cm 0.3cm 0.3cm}, clip, width=\linewidth]{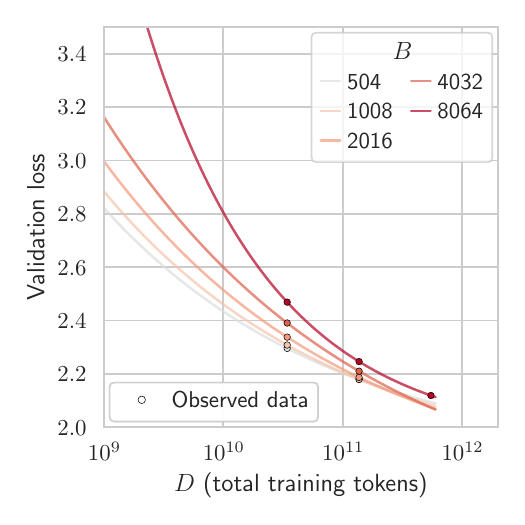}
    \caption{1.7B scale}
    \label{fig:quad_cbs_1_7B}
  \end{subfigure}
  \caption{\textbf{Scaling laws in $D$ for computing $\cbs$}: Full set
    of $B$-specific power laws, for 111M to 1.7B scales, fitted after
    training models with different batch sizes, $B$, and dataset
    sizes, $D$, at each scale (empirical data from real training runs
    indicated by points).  From these power laws, we can compute the
    amount of data needed to reach any target loss, as illustrated in
    main paper \cref{fig:cbs_fitting}.\label{fig:quad_cbs_fitting}}
\end{figure*}

\cref{fig:quad_cbs_fitting} illustrates all the fit scaling laws for
the $\cbs$ experiments.  Notice that beyond the fitting points, the
curves may not predict behavior well.
In particular, we would expect all curves to eventually converge as
$D$ increases.  Because loss targets beyond the fitting points are
unreliable, we only compute $\cbs$ at loss targets where all the data
sizes can be estimated through interpolation.

\begin{figure*}[th]
  \centering
  \begin{subfigure}[b]{0.49\textwidth}
    \centering
    \includegraphics[trim={0.3cm 0.4cm 0.3cm 0.3cm}, clip, width=\linewidth]{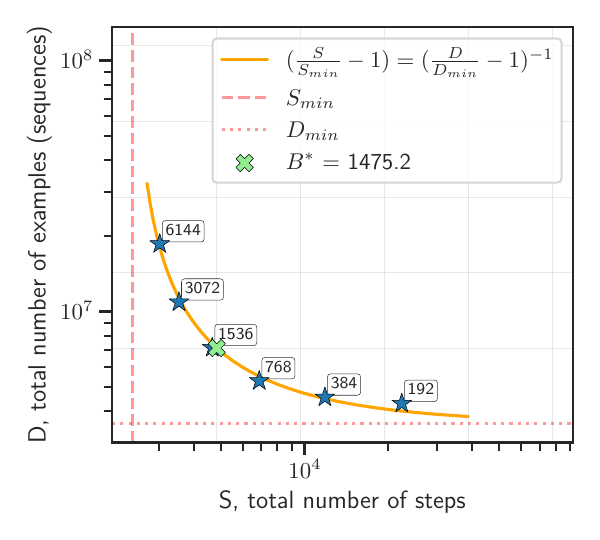}
    \caption{111M, loss=2.872}
    \label{fig:quad_cbs_fits_111M}
  \end{subfigure}
  \hfill
  \begin{subfigure}[b]{0.49\textwidth}
    \centering
    \includegraphics[trim={0.3cm 0.4cm 0.3cm 0.3cm}, clip, width=\linewidth]{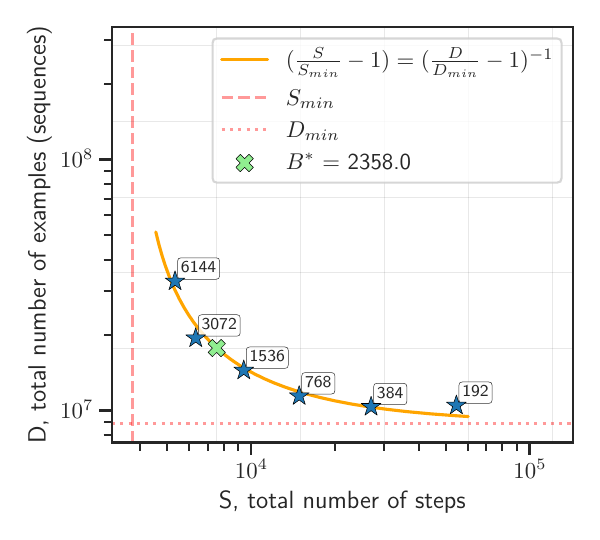}
    \caption{266M, loss=2.600}
    \label{fig:quad_cbs_fits_266M}
  \end{subfigure}
  % Next row
  \begin{subfigure}[b]{0.49\textwidth}
    \centering
    \includegraphics[trim={0.3cm 0.4cm 0.3cm 0.3cm}, clip, width=\linewidth]{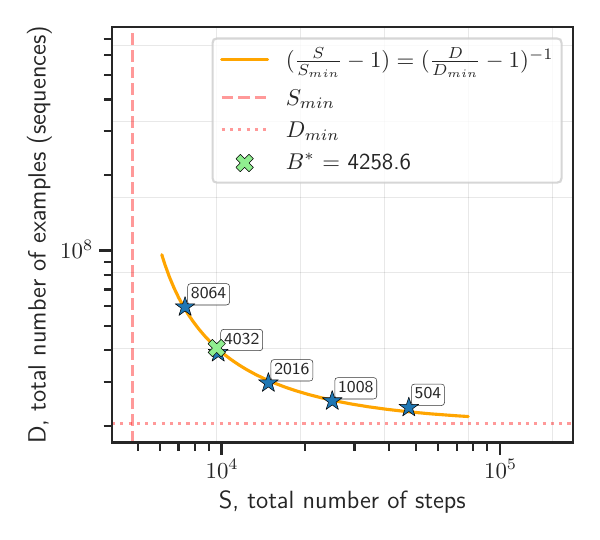}
    \caption{610M scale, loss=2.437}
    \label{fig:quad_cbs_fits_617M}
  \end{subfigure}
  \hfill
  \begin{subfigure}[b]{0.49\textwidth}
    \centering
    \includegraphics[trim={0.3cm 0.4cm 0.3cm 0.3cm}, clip, width=\linewidth]{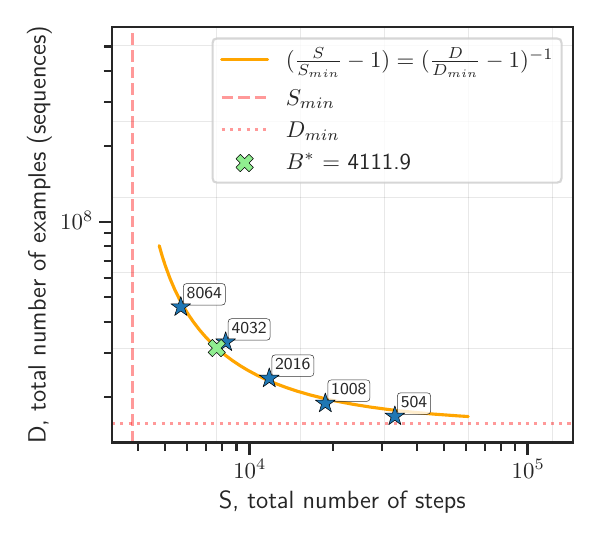}
    \caption{1.7B scale, loss=2.295}
    \label{fig:quad_cbs_fits_1_7B}
  \end{subfigure}
  \caption{\textbf{Example fits of trade-off \cref{eq:tradeoff}
      plotting $\smin$ and $\dmin$}: The empirical data has a good fit
    with \cref{eq:tradeoff} across model scales and loss targets.
    These estimates of $\cbs$ are used in fitting the $\cbs$-scaling
    power law
    (\cref{fig:hook} $\rightfig$).\label{fig:quad_cbs_fits}}
\end{figure*}

\cref{fig:quad_cbs_fits} shows the specific fits of \cref{eq:tradeoff}
at particular loss targets.  While \cref{eq:tradeoff} reflects the
data trend well over the given $B$ values, we consistently find that
points with very small $B$ do not approach $\dmin$.  We discussed this
observation further in \cref{sec:limitations}.

\renewcommand{\algorithmiccomment}[1]{$\triangleright$ #1}
\begin{algorithm}[tb]
  \caption{Generating the $\cbs$ power law}
  \label{alg:cbs_fitting}
  \begin{algorithmic}
    \small
    \STATE Initialize $\cbsPoints = [$ $]$
    \FOR{$N$ {\bfseries in} $\modelScales$}  % N
      \STATE \COMMENT{Fit $B$-specific (and $N$-specific) scaling laws, $L_{B}(D)$:}
      \FOR{$B$ {\bfseries in} $\batchSizes[N]$}  % B
        \STATE Reset $\scaleSet = [$ $]$
        \FOR{$D$ {\bfseries in} $10N, 20N, 80N, 320N, ...$} % D
          \STATE Train $\LLM(N,D,B)$, get validation loss $L'$
          \STATE $\scaleSet.add( \langle D,L' \rangle )$
        \ENDFOR % D
      \STATE Fit $\irr$, $\dconst$, $\beta$ for $L_{B}(D) = \irr + \dconst D^{-\beta}$ on $\scaleSet$
      \ENDFOR % B
      \STATE \COMMENT{Use $L_{B}(D)$ to estimate $\cbs$ at various loss values:}
      \FOR{$\hat{L}$ in $lossTargets[N]$} % L
        \STATE Reset $\tradeoffSet = [$ $]$
        \FOR{$B$ {\bfseries in} $\batchSizes[N]$} % B
          \STATE Get $\dinfer = L_{B}^{-1}(\hat{L}) = \left(\frac{\dconst}{\hat{L}-\irr}\right)^{\frac{1}{\beta}}$
          \STATE $\tradeoffSet.add( \langle \dinfer, S$=$\nicefrac{\dinfer}{B} \rangle )$
        \ENDFOR % B
        \STATE Fit $\dmin$, $\smin$ for \cref{eq:tradeoff} on $\tradeoffSet$
        \STATE $\cbs = \nicefrac{\dmin}{\smin}$
        \STATE $\cbsPoints.add( \langle \dmin, \cbs \rangle )$
      \ENDFOR % L
    \ENDFOR % N
    \STATE Fit $\ccbsalg$, $\mcbsalg$ for $\cbs = {\ccbsalg} (\dmin)^{\mcbsalg}$ on $\cbsPoints$
  \end{algorithmic}
\end{algorithm}

Finally, for clarity, we provide \cref{alg:cbs_fitting}, which gives
the detailed approach to generating the $\cbs$ power law in procedural
form.

\subsection{Estimating $\cbs$ for the 3.3B model}

To obtain an estimate of $\bcrit$ for the 3.3B model (shown in
\cref{fig:hook}, $\rightfig$), it was not feasible to apply our full
$\cbs$ fitting procedure at this scale (i.e., fitting $B$-specific
loss power-laws, etc.).
Instead, we estimated $\bcrit$ based on two $B$, $\dmin$ pairs.
That is, (based on an initial estimate of $\bcrit$) we trained a 3.3B
model to 23TPP with $B$=2016 and got a loss of 2.1688, and a separate
model to 30TPP with $B$=4032, obtaining a loss of 2.1695.
Given these losses are very close, these two models should have the
same $\bcrit$ and thus same $\dmin$.

We solve for this $\bcrit$ as follows.  Let $B_2 = 4032$ and $B_1 = 2016$.  By \cref{eq:s_extra}:
\begin{align*}
  D_2 &= \dmin(1 + B_2/\bcrit), \text{and} \\
  D_1 &= \dmin(1 + B_1/\bcrit)
\end{align*}
Let $r = D_2/D_1$ (i.e., 30/23 in this case).  If we divide the above
equations, and solve for $\bcrit$, we find:
\begin{align*}
  \bcrit &= \frac{B_2 - r B_1}{r - 1}
\end{align*}
Plugging in our values of $r$, $B_1$, and $B_2$, we obtain an
estimated $\bcrit$ of 4610, corresponding to a $\dmin$ of
approximately 16 TPP\@.

\subsection{$\cbs$ scaling in loss}\label{sec:appendix_loss_scaling}

\cref{fig:related} ($\middlefig$) shows that $\cbs$ is clearly not a
power law in loss, as proposed in prior
work~\citep{mccandlish2018empirical,kaplan2020scaling,li2024surge}.
However, if we only consider points with the same TPP, there does
appear to be somewhat of a power-law relationship.  In fact, this is
implied by $\cbs$ being a power law in $D$, along with the (standard)
assumption that loss scales similarly in $N$ and $D$.

Specifically, let $\hatr = {D}/{N}$ be the fixed TPP ratio.
Therefore, $N = {D}/{\hatr}$.  Assuming loss follows
\cref{eq:chinchilla}, we have:
\begin{align*}
L(N,D) &= E + \nconst N^{-\alpha} + \dconst D^{-\beta} \\
       &= E + \nconst \left( \frac{D}{\hatr} \right)^{-\alpha} + \dconst D^{-\beta} \\
       &= E + \nconst \hatr^{\alpha} D^{-\alpha} + \dconst D^{-\beta}
\end{align*}
Now, if $\alpha \approx \beta$, as is commonly
accepted~\citep{hoffmann2022empirical,besiroglu2024chinchilla,krajewski2024scaling,porian2024resolving,gadre2024language}, we have:
\begin{align*}
  L(D) &= E + (\nconst \hatr^{\alpha} + \dconst) D^{-\alpha} \\
   &= E + \kconst D^{-\alpha}
\end{align*}
where $\kconst = \nconst \hatr^{\alpha} + \dconst$ is a constant.  In
other words, at a fixed TPP, loss is a power law in data.  Given
$\bcrit$ is also fundamentally a power law in data, then by the
transitivity of power law relationships, $\cbs$ is also a power law in
loss in this context.  This relationship can also be derived by
expressing the $D$ in \cref{eq:cbs_scaling} as a power law in $\cbs$,
and substituting into $E + K D^{-\alpha}$.

\subsection{Weight decay affects $B$ scaling laws}\label{sec:app_wd_affects_b}

\begin{table}[]
  \centering
  \caption{Different $\lambda$ settings systematically affect fitted
    power laws for $\bopt$, and result in poorer fit quality (lower
    $\Rtwo$).  Fitted parameters change with $\lambda$ and
    consequently projected $\bopt$ values (in sequences) for different
    token budgets $D$.\label{tab:wd_bopt}}
  \begin{tabular}{@{}llllll@{}}
    \toprule
    Weight decay          & Scaling law                & $\Rtwo$ & D=1e10          & D=1e11          & D=1e12          \\ \midrule
    0.4                   & $\bopt = 0.0006 D^{0.607}$ & 0.706  & 587             & 2377            & 9615            \\
    0.2                   & $\bopt = 0.0012 D^{0.543}$ & 0.926  & 323             & 1128            & 3937            \\
    0.1                   & $\bopt = 0.0123 D^{0.429}$ & 0.972  & 240             & 644             & 1729            \\
    0.05                  & $\bopt = 0.384 D^{0.270}$  & 0.689  & 192             & 358             & 667             \\
    0.025                 & $\bopt = 10.3 D^{0.120}$   & 0.161  & 163             & 215             & 284             \\
    \emph{Tuned}          & $\bopt = 0.0306 D^{0.383}$ & \emph{0.984} & \emph{207}      & \emph{500}      & \emph{1207}     \\ \bottomrule
  \end{tabular}
\end{table}

\paragraph{Weight decay affects scaling of $\bopt$.}

\cref{tab:wd_bopt} illustrates how fixing $\lambda$ at different
values alters the fitted $\bopt$ power law. Larger $\lambda$
systematically yields larger $\bopt$ and poorer fit quality
(lower~$\Rtwo$). When $\lambda$ is fixed, the batch size that
minimizes loss is only a \emph{conditional optimum} (we denote it as
$\boptcond$) because it compensates for suboptimal
timescales~$\tema$ rather than representing the globally
tuned $\bopt$ obtained when $\lambda$ is optimized jointly.

The degraded $\Rtwo$ values arise because a fixed $\lambda$ forces $B$
to balance two partially competing goals: (1)~maintaining a good
$\tema$ value, and (2)~remaining below $\bcrit$ to avoid gradient
redundancy. A simple power law cannot capture this coupled behavior.

From the relation $\tema = B/(\eta\lambda D)$ and the empirical
scaling $\temaopt \propto (\nicefrac{D}{N})^{m}$
(\cref{eq:tepoch_scaling}), one can derive that, for constant $\eta$,
$\lambda$, and $N$, the batch size preserving $\temaopt$ should scale
as $B \propto D^{m+1}$ (with $m+1\!\approx\!0.47$ for our data). The
exponents in \cref{tab:wd_bopt}, however, deviate from~0.47. For
small~$D$ and large~$\lambda$, the $B$ that preserves $\temaopt$ lies
near or above $\bcrit$, yielding worse loss due to gradient
redundancy.  This means $\boptcond$ is artificially lower for small
$D$ values. Since $\boptcond$ is affected differently for different
$D$, the scaling law slope is distorted (in this case,
increased). Analogous issues disrupt $\boptcond$ for small $\lambda$.
These distortions reduce~$\Rtwo$ and impair generalization to
large-scale training. When $\lambda$ is tuned, this confound is
removed, and the resulting $\bopt$ law aligns cleanly with the
expected~$D^{0.4}$ scaling.

\paragraph{Weight decay affects scaling of $\bcrit$.}

A similar confound arises for~$\bcrit$. Fixing $\lambda$ causes
deviations from $\temaopt$ to mingle with true gradient-redundancy
effects.
At 111M scale and a target loss of~3.03, e.g., larger batches perform
worse solely because $\tema$ drifts from its optimal value, reducing
the fitted $\bcrit$ from~867 (tuned~$\lambda$) to~707
(fixed~$\lambda{=}0.1$).
At other loss targets, $\bcrit$ is less affected. Since
the estimated $\bcrit$ is affected differently at different scales,
the slope of the $\bcrit$ scaling law is again artificially
distorted. $\bcrit$ will appear to increase faster than $D^{0.5}$, and
projections to larger scales will be inaccurate.
In contrast, tuning~$\lambda$ isolates gradient-redundancy effects,
yielding a stable~$D^{0.5}$ relation that generalizes across scales.

\end{document}